\pdfoutput=1
\documentclass[sigconf]{acmart}

\AtBeginDocument{%
  }

\usepackage{epstopdf}
\usepackage{booktabs}
\usepackage{multirow}
\usepackage{colortbl}
\usepackage{amsmath}
\usepackage[ruled,linesnumbered]{algorithm2e}
\usepackage{svg}
\usepackage{enumitem}
\AtEndPreamble{
    \usepackage[capitalize]{cleveref}
    \crefname{section}{Sec.}{Secs.}
    \Crefname{section}{Section}{Sections}
    \Crefname{table}{Table}{Tables}
    \crefname{table}{Tab.}{Tabs.}
}
\setcopyright{acmlicensed}
\copyrightyear{2024}
\acmYear{2024}
\acmDOI{10.1145/3664647.3681241}

\acmConference[MM '24]{Proceedings of the 32nd ACM International Conference on Multimedia}{October 28-November 1, 2024}{Melbourne, VIC, Australia}
\acmBooktitle{Proceedings of the 32nd ACM International Conference on Multimedia (MM '24), October 28-November 1, 2024, Melbourne, VIC, Australia}
\acmISBN{979-8-4007-0686-8/24/10}

\begin{document}

\title{GenUDC: High Quality 3D Mesh Generation with Unsigned Dual Contouring Representation}

\author{Ruowei Wang}
\orcid{0009-0003-9112-1712}
% \email{wangruowei1027@gmail.com}
\affiliation{%
  \institution{Sichuan University}
  \city{Chengdu}
  % \state{Sichuan}
  \country{China}
}
\email{wangruowei1027@gmail.com}

\author{Jiaqi Li}
\orcid{0009-0008-0826-3047}
\affiliation{%
  \institution{Sichuan University}
  \city{Chengdu}
  % \state{Sichuan}
  \country{China}
  }
\email{2023226040007@stu.scu.edu.cn}

\author{Dan Zeng}
\orcid{0000-0002-9036-7791}
\affiliation{%
  \institution{Southern University of Science and Technology }
  % \streetaddress{1 Th{\o}rv{\"a}ld Circle}
  \city{Shenzhen}
  \country{China}
}
\email{zengd@sustech.edu.cn}

\author{Xueqi Ma}
\orcid{0009-0004-0203-8501}
\affiliation{%
  \institution{Shenzhen University}
  % \streetaddress{1 Th{\o}rv{\"a}ld Circle}
  \city{Shenzhen}
  \country{China}}
\email{qixuemaa@gmail.com}

\author{Zixiang Xu}
\orcid{0009-0008-6672-004X}
\affiliation{%
  \institution{Sichuan University}
  % \streetaddress{1 Th{\o}rv{\"a}ld Circle}
  \city{Chengdu}
  \country{China}}
\email{xzx34@stu.scu.edu.cn}

\author{Jianwei Zhang}
\orcid{0000-0002-5491-1745}
\affiliation{%
  \institution{Sichuan University}
  % \streetaddress{1 Th{\o}rv{\"a}ld Circle}
  \city{Chengdu}
  \country{China}}
\email{zhangjianwei@scu.edu.cn}

\author{Qijun Zhao}
\authornote{Corresponding Author.}
\orcid{0000-0003-4651-7163}
\affiliation{%
  \institution{Sichuan University}
  % \streetaddress{1 Th{\o}rv{\"a}ld Circle}
  \city{Chengdu}
  \country{China}}
\email{qjzhao@scu.edu.cn}

\renewcommand{\shortauthors}{Ruowei Wang et al.}
%% No italics, no superscripts
%% Use footnote or author note to identify equal contribution and/or contact author info

% \begin{abstract}
%   A clear and well-documented \LaTeX\ document is presented as an
%   article formatted for publication by ACM in a conference proceedings
%   or journal publication. Based on the ``acmart'' document class, this
%   article presents and explains many of the common variations, as well
%   as many of the formatting elements an author may use in the
%   preparation of the documentation of their work.
% \end{abstract}
\begin{abstract}

% 2024.4.8版：形状生成的意义。当前mesh生成的缺陷（要么太平滑，要么太难。）对论文的高层次概括。实验怎么样。分享代码
% 多用主动语态
Generating high-quality meshes with complex structures and realistic surfaces is the primary goal of 3D generative models. 
Existing methods typically employ sequence data or deformable tetrahedral grids for mesh generation. 
However, sequence-based methods have difficulty producing complex structures with many faces due to memory limits. 
The deformable tetrahedral grid-based method MeshDiffusion fails to recover realistic surfaces due to the inherent ambiguity in deformable grids. 
We propose the GenUDC framework to address these challenges by leveraging the Unsigned Dual Contouring (UDC) as the mesh representation. 
UDC discretizes a mesh in a regular grid and divides it into the face and vertex parts, recovering both complex structures and fine details. 
As a result, the one-to-one mapping between UDC and mesh resolves the ambiguity problem. 
In addition, GenUDC adopts a two-stage, coarse-to-fine generative process for 3D mesh generation. 
It first generates the face part as a rough shape and then the vertex part to craft a detailed shape. 
Extensive evaluations demonstrate the superiority of UDC as a mesh representation and the favorable performance of GenUDC in mesh generation. 
The code and trained models are available at \href{https://github.com/TrepangCat/GenUDC}{https://github.com/TrepangCat/GenUDC}.

\end{abstract}



\begin{teaserfigure}
  \centering
  % \fbox{\rule{0pt}{2in} \rule{\textwidth}{0pt}}
  \includegraphics[width=0.99\textwidth]{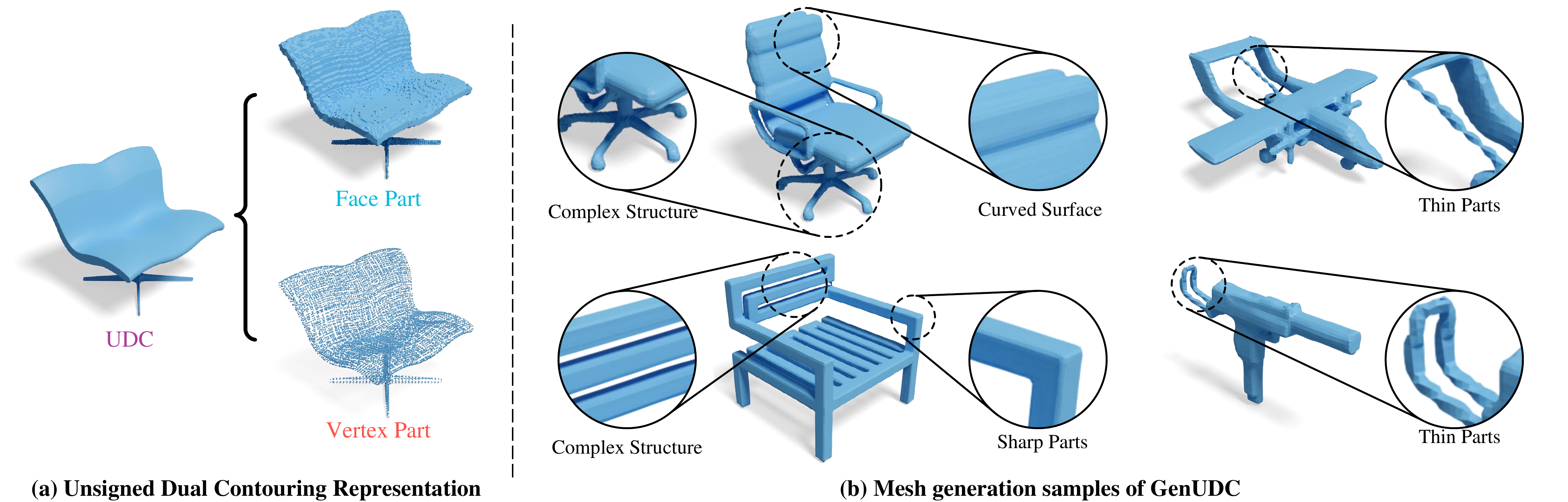}
  \caption{
  (a) A visual sample of Unsigned Dual Contouring Representation (UDC) consisting of the face part and vertex part. 
  (b) Our high-quality mesh generation results in $64^3$ resolution with close-up views.
  }
  \Description{The figure on the first page.}
  \label{fig: first}
\end{teaserfigure}


\maketitle

\section{Introduction}
\label{sec: intro}

% 1. 研究背景：简要介绍研究领域的背景和重要性，为读者提供必要的上下文。介绍该问题需要解决的一些关键问题。
%    具体内容：
Mesh plays an important role in 3D content generation and reconstruction \cite{traverser, sketch2mesh, fully-understanding-generic-objects-modeling-segmentation-and-reconstruction,2d-gans-meet-unsupervised-single-view-3D-reconstruction}, AR/VR \cite{ar_vr}, robotics \cite{robotics1, robotics2}, and autonomous driving \cite{autodrive1, autodrive2, autodrive3} and other 3D tasks \cite{3d_face_lizhifeng, brl_1_bird, traverser, brl_3_face}. 
It can flexibly represent various complex geometric shapes. 
High editability allows meshes to be modified and adjusted easily in computer-aided design (CAD). 
Additionally, it is effortless for users to convert meshes to other 3D representations, e.g., voxels, point clouds, and neural implicit functions. 
Besides, the rendering pipelines are designed for meshes, enabling high-quality 3D visualization effects.
However, employing deep neural networks on meshes is tricky because the numbers of vertices and faces are constantly changing, and modeling the complex topology structure of faces is also an obstacle. 
To navigate those challenges, a mesh representation compatible with deep learning and a capable generative framework adapted to this mesh representation are both highly desired.

% 2. 相关工作：概述领域内的相关研究和已有工作，指出当前研究的位置和研究空白。
Most existing approaches focus on intermediate representations, e.g., voxels \cite{3D-gan, CLIP-Sculptor}, point clouds \cite{pc-gan, luo2021diffusion, slide, brl_2_face}, neural implicit functions \cite{im_gan, sdfstylegan, SALAD, lasdiffusion, traverser} and so on, which are highly compatible with deep learning. 
However, those methods require a post-processing step \cite{dc, ndc, manifoldDC, lorensen1987marchingcubes, chernyaev1995marchingcubes33, nmc} to extract meshes, resulting in over-smooth surfaces and lacking detailed geometry. 
PolyGen \cite{polygen} first treats vertices and faces as sequences and uses transformer networks \cite{transformer} to generate vertices and then faces. 
MeshGPT \cite{meshgpt} and PolyDiff \cite{polydiff} follow similar ideas but concentrate on faces. 
All three approaches cannot produce mesh with intricate geometry since the memory limits the number of faces to no more than 2800.
MeshDiffusion \cite{meshdiffusion} chooses to combine a deformable tetrahedral grid with Signed Distance Functions (SDF) to model meshes. 
However, its data preparation is especially slow (\cref{table: data_fitting}), and the generated meshes are crumpled due to the deformable nature of the grid and the inaccurate 2D image supervision.

% 3. 本文的宏观解决思路。
To employ deep neural networks on meshes and synthesize meshes of high quality with complex structures, we construct a novel framework dubbed \textbf{GenUDC} to combine the Unsigned Dual Contouring representation (UDC) with a two-stage, coarse-to-fine generative process, enabling high-quality mesh generation.
As a mesh counterpart, UDC consists of a face part and a vertex part.
Accordingly, we decompose the mesh generation into two subtasks: the face part generation and vertex part generation.
To address these subtasks, we devise a customized pipeline, which involves generating faces first and then vertices.

% 4. 具体的方法和途径：
% 新版本
Precisely, to find a proper mesh representation, we expand Dual Contouring \cite{dc}, which has long been regarded as an isosurface reconstruction method, to generation tasks.
Thus, we obtain the UDC representation to model meshes as shown in \cref{fig: first} (a) and \cref{fig: framework}.
In UDC, we discretize a mesh in a regular grid. 
The faces part of UDC is a set of tiny faces represented by boolean values.
The vertex part of UDC contains all the actual and potential vertices of those tiny faces.
Since the values of the face part and vertex part are arranged in a regular grid, we can conveniently employ deep learning-based generative models to learn the distribution of UDC.

Another pivotal component is the two-stage, coarse-to-fine generative process specially designed for UDC. 
Because the mesh is discretized in UDC, the face part draws the rough shape, and the vertex part describes the details. 
Consequently, we first employ a latent diffusion model to generate the face part, determining the mesh's rough shape and topological structure. 
Then, conditioning the rough shape, we take a vertex refiner to generate the vertex part.
Such a pipeline is a natural solution for mesh generation.
Without this pipeline, the edges would be jagged due to the inaccurate vertex part.
We will study the necessity of this pipeline in \cref{sec: ablation}.
% \TODO{}

% 旧版本
% To address these challenges, we construct a novel framework dubbed \textbf{GenUDC} to apply a discretized counterpart of mesh on generative models, enabling high-quality mesh generation. 
% To first craft a discretized mesh representation, we expand Dual Contouring \cite{dc}, which has long been regarded as an isosurface reconstruction method, to generation tasks. 
% Thus, we get the Unsigned Dual Contouring representation (UDC) to model meshes as shown in \cref{fig: first} (a) and \cref{fig: framework}. 
% Then, thanks to the regular grid of UDC, we can conveniently employ deep learning-based generative models to learn the distribution of UDC.

% A pivotal component is that we introduce a two-stage strategy to our generative model, which is specially designed for UDC.
% As shown in \cref{fig: first} (a) and \cref{sec: method_udc}, UDC consists of a vertex part $\mathcal{V}$ and a face part $\mathcal{F}$.
% So, we first employ a latent diffusion model to generate the face part, determining the rough shape and topological structure of the meshes. 
% Then, conditioning on the rough shape, we take a U-Net to generate the vertex part. Considering the tight relation between $\mathcal{F}$ and $\mathcal{V}$, this is a natural and efficient way for vertex part generation.
% If we ignore this relationship, it results in jagged edges due to the inconsistency between the faces and vertices as shown in \cref{fig: visual_withU_withoutU}.

Finally, using GenUDC, we can produce high-quality meshes with complex structures and realistic details as shown in \cref{fig: first} (b) and \cref{fig: visual}. 
Comprehensive experiments demonstrate our superiority over existing ones in mesh generation. 
In data fitting, compared with MeshDiffusion, our method runs at $3274\%$ times their speed and consumes only $13\%$ of their total memory as shown in \cref{table: data_fitting}.
% 然后介绍UDC的构成，针对这个构成，我们提出了什么生成办法

% 5. 关键贡献：强调论文的关键贡献和创新点，突出与已有研究的不同之处或改进之处。
To summarize, the contributions of this paper are:
\begin{itemize}[topsep=3pt]
\item 
We propose a novel framework, \textbf{GenUDC}, utilizing UDC as the representation for high-quality mesh generation.

\item We design a two-stage, coarse-to-fine generative pipeline to UDC, which generates faces and then vertices, circumventing the jagged edges problem.

% \item We propose to employ a U-Net to learn the relationship between faces and vertices in vertices generation, circumventing the jagged edges problem.

\item Extensive experiments demonstrate our method's superior performance in mesh generation and data fitting.
% 旧版
% \item Our method proves its superiority in mesh generation and data fitting, as proven by extensive experiments.
\end{itemize}

\section{Related Work}
\label{sec: relatedwork}

% Mesh plays a fundamental role in various fields, such as computer graphics, computer-aided design, computational physics, etc. 
% Over the past years, mesh has remained a hotspot of research. 
This section outlines some closely related topics to our study: 3D shape generation, isosurface reconstruction, and diffusion models.

\subsection{3D Shape Generation}
\label{sec: RW_shape_generation}

\begin{table}[t]
\centering

\caption{
Taxonomy of mesh generation methods.
% \TODO{}
% 是否需要后处理，face上上限，显存消耗，representation
}

\resizebox{0.8\linewidth}{!}{
\begin{tabular}{@{}l|ccc@{}}
\toprule
Method                             & Representation                                                                                   & Memory                      & Maximum Num Of Faces                  \\ \midrule
PolyGen \cite{polygen}             & {\color[HTML]{333333} \begin{tabular}[c]{@{}c@{}}Face Sequence\\ + Vertex Sequence\end{tabular}} & {\color[HTML]{333333} High} & {\color[HTML]{333333} Less Than 2800} \\ \midrule
MeshGPT \cite{meshgpt}             & {\color[HTML]{333333} Triangle Face Sequence}                                                    & {\color[HTML]{333333} High} & {\color[HTML]{333333} Less Than 800}  \\ \midrule
PolyDiff \cite{polydiff}           & {\color[HTML]{333333} Triangle Face Soup}                                                        & {\color[HTML]{333333} High} & {\color[HTML]{333333} Less Than 800}  \\ \midrule
MeshDiffusion \cite{meshdiffusion} & \begin{tabular}[c]{@{}c@{}}Deformable Tetrahedral \\ Grid + SDF\end{tabular}                     & Medium                      & More Than 32768                       \\ \midrule
\textbf{Ours (GenUDC)}             & \textbf{UDC (Regular Grid)}                                                                      & \textbf{Medium}             & \textbf{More Than 32768}              \\ \bottomrule
\end{tabular}
}

\label{table: related_work}
\end{table}

With the advent of deep learning, researchers have been exploring the generation of 3D voxels \cite{3D-gan, g2l-gan, pagenet, voxel_energy, CLIP-Sculptor} and point clouds \cite{pc-gan, spgan, tree-gan, progressive-pcgan, energy_pc, pdgn, mrgan, pointflow, dpf, softflow, luo2021diffusion, slide} using neural networks. However, voxels suffer from memory limits, and point clouds lack topology of shapes. 
Until the dawn of neural implicit function \cite{deepsdf, im_gan, occupancyNet}, the community finds it an excellent shape representation, which does not require a lot of memories and can be easily transformed into meshes. 
The neural implicit function is specially designed for advanced deep neural networks and inspires a lot of work \cite{grid-im-gan, sdfstylegan, shapegan, nfd, sdf_diffusion1, T2TD, fullformer, Li_2023_CVPR_diffusion_sdf, Chou_2023_ICCV_diffusion_sdf, 3DShape2VecSet, SALAD, 3dqd, SDFusion, lasdiffusion, TAPS3D, traverser, wavelet}. 
It utilizes SDF values or occupancy values as the intermediate representation of 3D shapes. 
By using some isosurface reconstruction methods like Marching Cube \cite{lorensen1987marchingcubes, chernyaev1995marchingcubes33}, and Dual Contouring \cite{dc}, meshes can be reconstructed from those neural implicit functions. 
However, this also means those implicit function-based methods still require a post-processing step and cannot directly generate meshes. 
% What is more, due to the continuity of a neural implicit function, it tends to produce over-smooth meshes as shown in \cref{fig: visual}.  这一句话的必要性并不足够。我觉得这一段的要点，在于描述以前的隐表示、点云、voxel等方法需要后处理才能得到mesh。

% % DreamFusion系列的工作
% A collection of existing methods \cite{dreamfusion, magic3d, fantasia3d, prolificdreamer, dreamgaussian, gsgen, point_to_3d, control3d, 3d_fingertips} adapt Neural Radiance Fields (NeRF) or 3D Gaussian Splatting (3DGS) as 3D representation. 
% They utilize the powerful text-to-image generative model, Stable Diffusion \cite{latent-diffusion}, as the guidance to optimize NeRF or 3DGS with a text prompt.
% After optimization, the final NeRF or 3DGS contains both 3D shape and texture information.
% The mesh can be extracted from it by some post-processing methods.
% However, they are time-consuming, taking hours of optimization for each text prompt. 
% They also suffer from artifacts such as over-saturated colors and the multi-face problem.

Moreover, some works are trying to find a proper mesh representation to generate meshes directly. 
PolyGen \cite{polygen}, MeshGPT \cite{meshgpt}, and PolyDiff \cite{polydiff} are inspired by natural language processing to process meshes as sequences. 
By leveraging the power of transformer network \cite{transformer}, they can theoretically produce vertices and faces of any length. 
In practice, the limited memory constrains the complexity of synthetic mesh structures, making it difficult for them to generate curved surfaces.
A similar work \cite{maxueqi2024eccv} has explored to generate wireframes of mesh.
% \textcolor{red}{to be edited. see the comments in the latex.} 
% Unconditional mesh generation has been a long-standing problem with two difficulties: (1) how to generate faces which are connections between vertices. (2) different meshes have different numbers of vertices. Recently, two works have tried to solve this problem. 
% PolyGen \cite{polygen} treats vertices and faces as sequential data. Using transformer networks \cite{transformer}, PolyGen first generates vertices one by one and then predicts faces by connecting existent vertices. Theoretically, It can produce vertices and faces of unlimited length. However, since transformer networks require too much memory, PolyGen has to limit meshes with no more than 800 vertices and 2800 faces. 
In MeshDiffusion \cite{meshdiffusion}, a deformable tetrahedral grid and SDF values are utilized to recover meshes. 
It supposes all mesh vertices are on the edges of the deformable tetrahedral grid. 
It can use linear interpolation to compute mesh vertices with the coordinates of adjacent grid points and SDF values. 
After getting the mesh vertices, It produces faces by connecting mesh vertices in the same tetrahedrons. 
However, the deformable grid brings ambiguity to the fitting mesh. 
Two different tetrahedral grids with distinct SDF values may recover the same mesh. 
In addition, to fit a mesh, the deformable grid is trained in the supervision of rendered images, which are inaccurate due to various rendering settings. 
The ambiguity and 2D supervision tend to result in deficient surfaces shown in \cref{fig: data_fitting}. 
As for data preparation, it takes too much time and memory to fit a tetrahedral grid on a shape due to the 2D supervision, as shown in \cref{table: data_fitting}.

In contrast to sequence-based methods \cite{polygen, meshgpt, polydiff}, our method is capable of using limited memory to generate a diverse range of mesh structures, such as flat surfaces, thin parts, curved surfaces, sharp parts, and so on, as shown in \cref{fig: first} (b). 
Compared with MeshDiffusion \cite{meshdiffusion}, we use a regular grid to fit meshes with more accurate results, less processing time, and less memory, as shown in \cref{table: data_fitting} and \cref{fig: data_fitting}.
We taxonomize methods that can directly generate meshes in \cref{table: related_work}.
We present more details of the data fitting comparison between MeshDiffusion and ours in \cref{sec: data_fiting}.

\subsection{Isosurface Reconstruction}
Typically, isosurface reconstruction methods extract meshes from volume data (e.g. voxels and SDF volumes). 
As a pioneering work, the original Marching Cubes (MC) method is proposed by Lorensen and Cline \cite{lorensen1987marchingcubes}. 
It discretizes a mesh into a regular grid and creates approximative surfaces in each cube according to intersections between the mesh and grid. 
Its most well-known variant, MC33 \cite{chernyaev1995marchingcubes33}, can even model all possible topological cases in a cube. 
However, since vertices of approximative surfaces are on the edges of the grid, it is hard for the marching cubes method to model sharp parts. 
The Dual Contouring method (DC) \cite{dc} is thus proposed. 
Its vertices of approximative faces (also called dual faces) are in the cubes. 
So DC can recover sharp parts. 
With the rise of deep learning, Deep Marching Cubes \cite{deepmc} first applied deep learning to isosurface reconstruction. 
Neural Marching Cubes (NMC) \cite{nmc} and Neural Dual Contouring (NDC) \cite{ndc} focus on building a learnable version of MC and DC. 
Manifold Dual Contouring \cite{manifoldDC} and FlexiCubes \cite{flexicubes} try to solve the non-manifold problem in DC.
VoroMesh \cite{VoroMesh} introduces Voronoi diagrams to isosurface reconstruction.

All isosurface reconstruction methods focus on transforming 3D data of various forms into mesh counterparts. 
In contrast, we adopt UDC and expand it to shape generation by learning the distribution of the UDC representations. 
In other sections, with a little abuse of the abbreviation, we refer to UDC representation as UDC.

% 讨论这部分的时候，可以提一下重建和生成本质上都是一种生成任务，重建是一种条件生成，我们这里的mesh generation是一种无条件的生成。

\subsection{Diffusion Models}
Diffusion models are a class of deep generative models that play an important role in artificial intelligence generated content (AIGC). 
It achieves pleasant results in various applications, such as image generation \cite{diffusion-model, latent-diffusion}, shape generation \cite{meshdiffusion}, text-to-3D \cite{Li_2023_CVPR_diffusion_sdf, Chou_2023_ICCV_diffusion_sdf}, etc.
Diffusion models are designed to model the step-by-step transformations between a simple distribution (e.g. Gaussian distribution) and a complex distribution of data.
Once trained, a diffusion model can map a sample of the simple distribution to the desired data distribution.
As a milestone of diffusion models, the Denoising Diffusion Probabilistic Model (DDPM) \cite{diffusion-model} introduces variational inference into diffusion models and shows greater potential over generative adversarial networks \cite{gan_origin}. 
But it still suffers from the huge memory requirement. 
Therefore, the latent diffusion model (LDM) \cite{latent-diffusion} proposes to train diffusion models in a low-dimensional latent space instead of the high-dimensional data space.
It has been demonstrated that this technique can speed up training and reduce memory footprints without degradation of generation quality.
In this paper, we adopt the LDM in the face part generation (see \cref{sec: method_face_gen}) since the regular grid takes a lot of memory footprints.

% 指出扩散模型属于哪一类、大概包含什么流程。有什么缺点，引入DDIM，然后引入Latent Diffusion，接着指出，我们为了什么使用Latent Diffusion.

\section{Method}

\subsection{Overview}
% 我们的目标是什么？ 需要解决什么问题？ 我们怎样解决了该问题？ 方法主要由那几部分构成，我们的方法为解决这个问题做出了何种贡献？ 本文接下来将如何组织？
How to represent meshes and process meshes with neural networks are two critical issues hindering mesh generation. 
To address them, we propose GenUDC, a novel generative framework for mesh generation. 
In GenUDC, we discretize a mesh in a regular grid to get its corresponding Unsigned Dual Contouring representation (UDC). 
Thus, due to UDC's regular grid structure, neural networks can easily be used on both watertight and non-watertight meshes. 
We further propose a two-stage, coarse-to-fine pipeline adapted to UDC, which generates faces and vertices successively. 
In summary, we offer a new and straightforward solution for mesh generation. 

In the following sections, we first elaborate on UDC in \cref{sec: method_udc}. 
Then, we illustrate our generative models for face generation in \cref{sec: method_face_gen} and vertex generation in \cref{sec: method_vertex_gen}. 
Finally, the implementation details are presented in \cref{sec: method_imple}.

\begin{figure*}[t]
    \centering
    \includegraphics[width=0.9\linewidth]{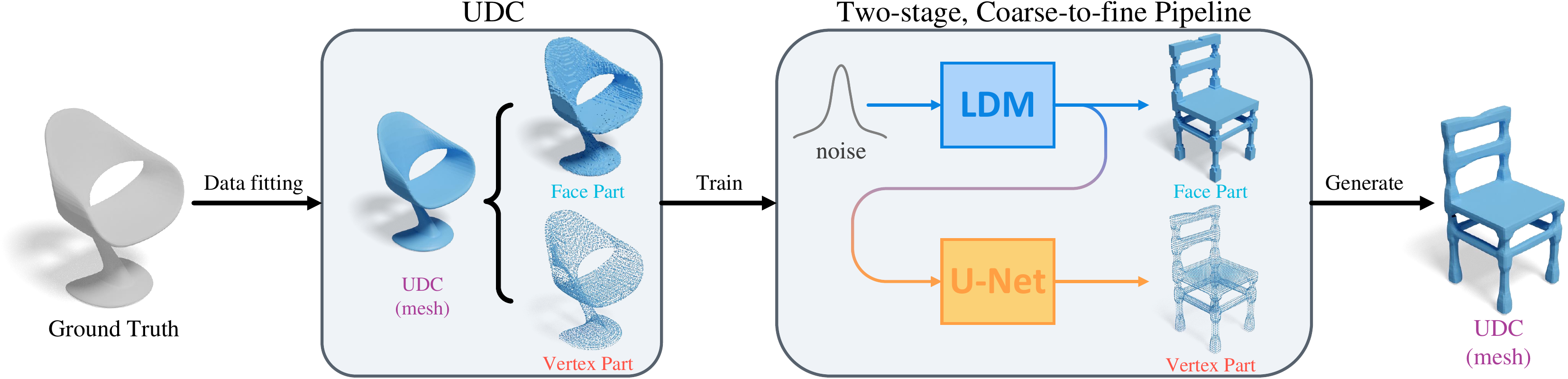}

    \caption{
    The overview of GenUDC. 
    It consists of UDC and a two-stage, coarse-to-fine generative pipeline. 
    We first translate the meshes to UDCs by data fitting.
    Then, we take UDCs to train the generative models.
    After training, we can generate the face part and vertex part to compose the output UDC.
    }
    % 需要包含的要素，主要着重描绘我们自己的东西：
    % 1. UDC是什么构成？ 以及其对应的可视化。
    % 2. GenUDC是整个网络框架，包含UDC和网络模型的部分。
    % 3. 所谓的两阶段是怎样的？
    \label{fig: framework}
\end{figure*}

\subsection{Unsigned Dual Contouring Representation}
\label{sec: method_udc}

We have briefly shown the main ideas of the Unsigned Dual Contouring representation (UDC) in \cref{fig: first} (a) and \cref{fig: framework}. 
For more details, in a grid $\mathcal{G}$ with the size of $(X+1, Y+1, Z+1)$, UDC can be formalized as: 
\begin{equation}
\label{eq:udc}
\mathrm{UDC} = \left\{
\begin{matrix}
\begin{aligned}
 &\mathcal{V} \in \mathbb{R}^{3\times |\mathcal{C}|}, \quad \quad &\text{(vertex part)} \\  
 &\mathcal{F} \in \mathbb{B}^{|\mathcal{E}|}, \quad \quad &\text{(face part)}
\end{aligned}
\end{matrix}\right. 
\end{equation}
where $\mathcal{C}$ are the cubes in the grid, $\mathcal{V}$ are the vertices, $\mathcal{E}$ are the edges inside the grid, and $\mathcal{F}$ are the faces (also called dual faces) denoted by the intersection flags of edges. 
The grid $\mathcal{G}$ contains $(X+1)(Y+1)(Z+1)$ nodes. 
There are $|\mathcal{C}|=XYZ$ cubes in the grid, and each cube contains a vertex $v \in \mathcal{V}$. 
Considering the edges along the x-axis, y-axis, and z-axis, there are $|\mathcal{E}|=X(Y-1)(Z-1)+(X-1)Y(Z-1) + (X-1)(Y-1)Z$ inside edges. 
If the intersection flag of an edge is true, four adjacent vertices make up two triangle faces that are $\mathit{dual}$ to the edge.
In other words, the edge intersects with one of the two triangle faces when the flag is true.
If not, there is no face intersecting with this edge. 
When translating a UDC to a correlative mesh, we craft faces by traversing all intersection flags in $\mathcal{F}$ and remove a subset of $\mathcal{V}$ which are not in these faces.
By this means, faces and remaining vertices comprise the final mesh.

% 讲述UDC的优势：非水密，刚性网格，结构简单方便用深度学习进行处理
% This means it solves two critical problems in mesh generation: (1) how to represent and generate faces and (2) how to handle varying numbers of vertices.
Compared with the traditional SDF-based methods \cite{sdfstylegan, meshdiffusion}, which usually generate over-smooth shapes, UDC can easily model the sharp parts as shown in \cref{fig: data_fitting}. 
In addition, the rigid grid used in UDC is suited for deep neural networks and can produce more realistic surfaces than the deformable grid of MeshDiffusion \cite{meshdiffusion}, which will be evaluated in \cref{sec: data_fiting}. 
Moreover, UDC has the potential to model non-watertight shapes shown in \cref{fig: data_fitting}. 

In practice, $\mathcal{V}$ are the relative coordinates in each cube, which means $0 \leq min(\mathcal{V}) \text{\ and\ } max(\mathcal{V}) \leq 1$ and $\mathcal{F}$ are boolean values.
When $X=Y=Z$, we pad $\mathcal{F}$ with zeros to the same size as $\mathcal{V}$. 
We call $\mathcal{V}$ as the vertex part and $\mathcal{F}$ as the face part.
 
\paragraph{Data Fitting} 
We follow a similar procedure of DC \cite{dc} to fit a mesh with UDC. 
Given a mesh $\mathcal{M}=(\mathcal{V^M}, \mathcal{F^M})$ and a grid $\mathcal{G}=(\mathcal{C}, \mathcal{E})$, we first find the crossing vertices $\mathcal{V^{E}}$ of the mesh $\mathcal{M}$ on the edges $\mathcal{E}$. 
Then, we compute the normals $\mathcal{N^E}$ of $\mathcal{M}$ at those crossing vertices. 
With $\mathcal{V^E}$ and $\mathcal{N^E}$, we can create UDC as:
\begin{align}
\label{eq:mesh2udc_1}
&\mathcal{V}=f_{\mathcal{V}}(\mathcal{V^E}, \mathcal{N^E}),
\\
\label{eq:mesh2udc_2}
&\mathcal{F} = f_{\mathcal{F}}(\mathcal{V^E}, \mathcal{E}).
\end{align}

The dual contouring vertices $\mathcal{V}$ should be on the surfaces of $\mathcal{M}$. 
So we extrapolate neighboring normals $\mathcal{N^E}$ to find a point of best fit in each cube:
\begin{equation}
\label{eq:v}
f_{\mathcal{V}}: \{v_{xyz} | \underset{v_{xyz}}{\operatorname{arg\,min}}\sum_{e\in \mathcal{C}_{xyz}}^{}(\mathcal{N^{E}}_{e}\cdot(v_{xyz}-\mathcal{V^{E}}_{e}))^2\},
\end{equation}
where $v_{xyz}$ is the vertex inside the cube $\mathcal{C}_{xyz}$ which is indexed by $(x, y, z)$, and $e$ are 12 edges of $\mathcal{C}_{xyz}$. $0 \leq x < X$, $0 \leq y < Y$, and $0 \leq z < Z$. 
By default, if there is no $\mathcal{V^{E}}_{e}$ or $\mathcal{N^{E}}_{e}$ in a cube, $v_{xyz}$ is set to $[0.5, 0.5, 0.5]$.

Besides, we only craft faces $\mathcal{F}$ when $\mathcal{M}$ intersects with an edge $e \in \mathcal{E}$ at the crossing vertex $v \in \mathcal{V^E}$:
\begin{equation}
\label{eq:f}
f_{\mathcal{F}}: \left\{
\begin{matrix}
\begin{aligned}
 &1, \quad \text{if} \ \forall \ e \in \mathcal{E}, \ \exists \ v \in \mathcal{V^E} \ \text{is on the} \ e, \\  
 &0, \quad \text{otherwise}.
\end{aligned}
\end{matrix}\right.
\end{equation}

\subsection{Face Part Generation}
\label{sec: method_face_gen}

% 从处理点云、网格等不同形式的生成模型说起，一句话转到我们自己的方案上。说明UDC与我们自己的方案的适配性。为了三维数据格式的显存占用问题，我们选择了LDM。
% 中间这里再想想怎么写。
% 接着，写VAE，记得写清楚trick，以及该模型类似什么。然后写Diffusion model。之后，写training是怎么样的， 最后写inference要做什么。
In UDC, we have devised a simple and intuitive method for generating faces by connecting them with intersection flags. 
If an edge's intersection flag is true, it crosses faces. 
If not, there is no face. 
By this means, we can denote all faces of the mesh as boolean values and arrange them into a regular grid as a face tensor $\mathcal{F} \in \mathbb{B}^{|\mathcal{E}|}$. 
Thus, we can easily employ neural networks to face part generation.

% 为了节省显存，我们用了隐模型，隐模型有什么优点
To reduce the memory footprint, we use an LDM \cite{latent-diffusion} to learn the distribution of $\mathcal{F}$. 
Our LDM consists of a Variational AutoEncoder (VAE) \cite{vae} and a diffusion model \cite{latent-diffusion, diffusion-model}. 
VAE compresses a $\mathcal{F}$ to a latent representation $z$. 
Then, we take latent representations $z$ to train our diffusion model. 
Thus, by extracting the compression process from the generative learning phase, we can speed up the diffusion model training process and reduce the memory footprints. 
And since the latent space is perceptually equivalent to the input space, there is no quality reduction for the diffusion model. 
We provide detailed descriptions of VAE and the diffusion model below.

% 首先VAE干什么（写个公式），然后扩散模型干什么，为什么我们要用这个trick（因为值域），可以放到最后解释
\paragraph{VAE}
A VAE comprises an encoder $E$ and a decoder $D$.
Given $\mathcal{F} \in \mathbb{B}^{|\mathcal{E}|}$, we first normalize $\mathcal{F}$ to $[-1.0, 1.0]$ using min-max normalization.
Then $E$ encodes $\mathcal{F}$ to a mean code $\mu \in \mathbb{R}^{c\times d\times h\times w}$ and a standard deviation code $\sigma \in \mathbb{R}^{c\times d\times h\times w}$. 
We use the mean code $\mu$ as the latent code $z \in \mathbb{R}^{c\times d\times h\times w}$ without reparameterization, which differs from the typical VAE. 
Finally, $D$ decodes $z$ back to the face tensor $\mathcal{F}_{pred}=D(z)$.
% 写VAE，记得写清楚trick，以及该模型类似什么。
We train our VAE with the mean squared error (MSE) loss and the Kullback–Leibler divergence (KL) loss:
\begin{equation}
\label{eq:vae_loss}
    % \begin{aligned}
        \mathcal{L}_{vae}=\mathcal{L}_{mse}(D(E(\mathcal{F}))), \mathcal{F})
         + KL(\mathcal{N}(\mu, \sigma)||\mathcal{N}(0, 1)).
    % \end{aligned}
\end{equation}
Since we do not use the reparameterization technique, our VAE is more like an autoencoder (AE) producing compact latent codes (close to zero).

\paragraph{Diffusion Model}
After encoding the face part $\mathcal{F}$ to the latent code $z$ with our VAE, we employ a diffusion model \cite{diffusion-model, latent-diffusion} to the latent code distribution $p(z)$. 
We first normalize $z_0 \in p(z)$ to $[-1.0, 1.0]$. 
Then, through a series of diffusion steps, we introduce the controlled Gaussian noise $\epsilon$ to $z_0$ and transform it to $z_t=\sqrt{\bar{\alpha}_t}z_0+\sqrt{1-\bar{\alpha}_t}\boldsymbol{\epsilon}$, where $t=1\ldots T$ and $\bar{\alpha}_t=\prod_{i=1}^{t}\alpha_i$. 
$\alpha_t=1-\beta_t$ and $\beta_t$ is the predefined variance.
The diffusion model $\theta$ is trained to predict the noise $\epsilon$, aiming at reversing the diffusion steps. 
The training objective is
\begin{align}
\mathcal{L}_{dm}=  \mathbb{E}_{x,t,\epsilon \sim \mathcal{N}(0,1)} ||\boldsymbol{\epsilon} - \boldsymbol{\epsilon_\theta}(z_t, t)||_1 .
\end{align}
% Once trained, we can utilize the diffusion model to predict a denoised variant of $z_t$ and synthesize new data $z_0$.
After training, to generate a face part $\mathcal{F}$, a sampled Gaussian noise $\epsilon \sim \mathcal{N}(0,1)$ is seen as $z_T$. 
Then, our trained diffusion model denoises $z_T$ to $z_0$. 
$z_0$ is further denormalized from $[-1.0, 1.0]$ to the original data range of $p(z)$. 
Finally, the $D$ decodes denormalized $z_0$ to $\mathcal{F}$.

More details of the network are in the supplemental material.
% We provide more details of the network in the supplemental material.

% 可以先说一下对于VAE得到的compact code 我们做了什么预处理方法。然后是利用前向传递过程进行训练，使用了什么loss（可以找个合适的机会，去提一下模型是Unet，这里写的方式可以参考LDM）。这里可以只给loss，其他的公式都放到补充材料。
% 接着，说怎么做inference。

\subsection{Vertex Part Generation}
\label{sec: method_vertex_gen}

The vertex part $\mathcal{V}$ is a set of relative vertex coordinates, and all vertices are arranged in a regular grid. 
Each vertex is in a cube of this grid. 
The vertex part contains all actual and potential vertices of a mesh. 
Since several vertices compose a face, there is a tight correlation between $\mathcal{V}$ and $\mathcal{F}$.
% The face part $\mathcal{F}$ corresponds to a mesh's faces. 
% Since the vertices and faces of a mesh are highly correlated, there is also a tight connection between $\mathcal{V}$ and $\mathcal{F}$. 
Therefore, learning this correlation is a crucial problem in the vertex part generation.

% 一旦face部分被确定了，face部分会怎么样
In UDC, when the face part is determined, the rough shape is known, and the variance of the vertex part is limited. 
So we treat the vertex part generation as a regression task.
We take a vertex refiner to generate $\mathcal{V}$ conditioned on $\mathcal{F}$.
Here, we use a 3D version of U-Net \cite{unet} as the vertex refiner.
Note that $\mathcal{F}$ is padded to the same size as $\mathcal{V}$ described in \cref{sec: method_udc}.
Firstly, we normalize the face part $\mathcal{F}$ and the vertex part $\mathcal{V}$ to $[-1.0, 1.0]$. 
Secondly, the vertex refiner takes the $\mathcal{F}$ as the conditional input and generates a vertex part $\mathcal{V}_{pred} \in \mathbb{R}^{3\times |\mathcal{C}|}$ as shown in \cref{fig: framework}.
We compare $\mathcal{V}_{pred}$ with the ground truth $\mathcal{V}_{gt}$ to train networks:
\begin{align}
\mathcal{L}_{float}&=\mathcal{L}_{mse}(\mathcal{V}_{gt}, \mathcal{V}_{pred}),\\
\mathcal{V}_{pred}&=\mathrm{Unet3D}(\mathcal{F}),
\end{align}
where $\mathcal{L}_{mse}$ is MSE loss. 
$\mathcal{V}_{gt}$ is the ground truth vertices paired with $\mathcal{F}$. 
In the inference phase, $\mathcal{V}_{pred}$ is denormalized from $[-1.0, 1.0]$ to $[0.0, 1.0]$.

This is a natural and efficient solution to learn the correlation between $\mathcal{F}$ and $\mathcal{V}$ for vertex part generation with reasonable training costs. 
If we eliminate the vertex refiner and generate $\mathcal{F}$ and $\mathcal{V}$ together, synthesized meshes will contain jagged edges due to inaccurate vertex coordinates as discussed in \cref{sec: ablation}. 
% We will illustrate the necessity of the vertex refiner in \cref{sec: ablation}.

More details of the network are in the supplemental material.
% We provide more details of the network in the supplemental material.

% 写作梗概：同样得益于regular grid，我们能直接使用各种神经网络。
% \TODO{rewrite. } % 拓扑结构已经确定的情况下，用这种回归式的方法，更好，无论是效果上，还是速度上。

% We propose using U-Net to learn how to make this connection. In essence, the float-part generation is a bool-part conditioned generation task. So the U-Net takes a bool-part $\mathcal{F} \in \mathbb{B}^{3\times |\mathcal{C}|}$ normalized to $[-1.0, 1.0]$ as the input, and predicts its corresponding float-part $\mathcal{V}_{pred} \in \mathbb{R}^{3\times |\mathcal{C}|}$ as shown in \cref{fig: framework}. We train this U-Net with:
% \begin{align}
% \mathcal{L}_{float}&=\mathcal{L}_{mse}(\mathcal{V}_{gt}, \mathcal{V}_{pred}),\\
% \mathcal{V}_{pred}&=\mathrm{Unet}(\mathcal{F}),
% \end{align}
% where $\mathcal{L}_{mse}$ is MSE loss. $\mathcal{V}_{gt}$ is the ground truth vertices paired with $\mathcal{F}$.

% It is natural to learn the relationship between $\mathcal{V}$ and $\mathcal{F}$ using the U-Net. If we eliminate the U-Net in the float-part generation, generated meshes will contain jagged edges due to the inaccurate $\mathcal{V}$. We will illustrate the necessity of our network design in \cref{sec: ablation}.
% % 使用了什么loss。解释相关的符号。实践中，归一化到[-1,1]。解释这样做的意义，这是一个很直接natural的方式，用来学习bool-part与float-part之间的联系。后面关于这个，我们会在某个章节做相关的消融实验。
% We provide more details of the network in the supplemental material.

\subsection{Implementation Details}
\label{sec: method_imple}
If not specified otherwise, we set $X=Y=Z=64$ for the grid $\mathcal{G}$ and $c=64,\ d=h=w=16$ for the latent code $z$. 
During training, $\mathcal{V}$ and $\mathcal{F}$ are normalized to $[-1.0, 1.0]$. 
At the final step of mesh generation, we denormalize the generated $\mathcal{V}$ and $\mathcal{F}$ to $[0.0, 1.0]$ and keep $\mathcal{V}$ as floating-point numbers and $\mathcal{F}$ as boolean values. 
We train the VAE and U-Net with all five categories as told in \cref{sec: data_setting}. 
In contrast, the diffusion model is trained in a category-specific manner. 
We use the AdamW optimizer \cite{adamw} with $\beta_1,\ \beta_2=[0.9,\ 0.999]$ for all networks. 
Empirically, large $\beta$ values can make our diffusion model produce realistic meshes. 
During the inference of diffusion models, we adopt the sampling method in Denoising Diffusion Probabilistic Models \cite{diffusion-model} and set the inference step as 1000.

\section{Experiments}

% 如果缺字数，能够用这个来填充。
% We demonstrate the superiority of our generative model over the previous shape generation methods in \cref{sec: shape_generation}, compare our data fitting process with MeshDiffusion \cite{meshdiffusion} in \cref{sec: data_fiting}, and analyze the design of the vertex part generation in \cref{sec: ablation}.

\subsection{Data}
\label{sec: data_setting}
Following the protocol of MeshDiffuision \cite{meshdiffusion}, we use the ShapeNet Core (v1) dataset \cite{shapenet2015} to train and test our networks. 
Airplane, car, chair, refile, and table — five categories are used in our experiments. 
For each category, we split all data like \cite{hsp} and \cite{sdfstylegan} do: $70\%$ as the training set, $20\%$ as the test set, $10\%$ as the validation set.
To be clear, the validation set is not used. 
For a fair comparison, we remove the interior of shapes.
We apply the data-fitting method in \cref{sec: method_udc} to generate UDC for all mesh data.

\subsection{Shape Generation}
\label{sec: shape_generation}

\begin{table*}[t]
\centering

\caption{
Quantitative evaluation of shape generation in $64^3$ resolution. 
% 这里可以考虑Airplane-64和 Airplane单独列一行
}

\resizebox{0.76\linewidth}{!}{
\begin{tabular}{@{}c|l|cccccccccc@{}}
\toprule
\multirow{2}{*}{}         & \multirow{2}{*}{Method} & \multicolumn{3}{c}{MMD ($\downarrow$)}                                  & \multicolumn{3}{c}{COV ($\%, \uparrow$)}                              & \multicolumn{3}{c}{1-NNA ($\%, \downarrow$)}                          & \multirow{2}{*}{JSD $(10^{-3},\ \downarrow)$} \\ \cmidrule(lr){3-11}
                          &                         & CD $\times 10^3$ & EMD $\times 10$ & \multicolumn{1}{c|}{LFD}           & CD             & EMD            & \multicolumn{1}{c|}{LFD}            & CD             & EMD            & \multicolumn{1}{c|}{LFD}            &                                                 \\ \midrule
\multirow{4}{*}{Chair}    & IM-GAN                  & 13.928           & 1.816           & \multicolumn{1}{c|}{3615}          & \textbf{49.64} & 41.96          & \multicolumn{1}{c|}{47.79}          & 58.59          & 69.05          & \multicolumn{1}{c|}{68.58}          & 6.298                                           \\
                          & SDF-StyleGAN            & 15.763           & 1.839           & \multicolumn{1}{c|}{3730}          & 45.60          & 45.50          & \multicolumn{1}{c|}{43.95}          & 63.25          & 67.80          & \multicolumn{1}{c|}{67.66}          & 6.846                                           \\
                          & MeshDiffusion           & \textbf{13.212}  & 1.731           & \multicolumn{1}{c|}{3472}          & 46.00          & 46.71          & \multicolumn{1}{c|}{42.11}          & \textbf{53.69} & \textbf{57.63} & \multicolumn{1}{c|}{63.02}          & 5.038                                           \\
                          & \textbf{Ours}           & 14.083           & \textbf{1.653}  & \multicolumn{1}{c|}{\textbf{2924}} & 48.08          & \textbf{48.60} & \multicolumn{1}{c|}{\textbf{47.94}} & 59.18          & 58.67          & \multicolumn{1}{c|}{\textbf{60.84}} & \textbf{4.837}                                  \\ \midrule
\multirow{4}{*}{Car}      & IM-GAN                  & 5.209            & 1.197           & \multicolumn{1}{c|}{2645}          & 28.26          & 24.92          & \multicolumn{1}{c|}{30.73}          & 95.69          & 94.79          & \multicolumn{1}{c|}{89.30}          & 42.586                                          \\
                          & SDF-StyleGAN            & 5.064            & 1.152           & \multicolumn{1}{c|}{2623}          & 29.93          & 32.06          & \multicolumn{1}{c|}{41.93}          & 88.34          & 88.31          & \multicolumn{1}{c|}{84.13}          & 15.960                                          \\
                          & MeshDiffusion           & 4.972            & 1.196           & \multicolumn{1}{c|}{2477}          & 34.07          & 25.85          & \multicolumn{1}{c|}{37.53}          & 81.43          & 87.84          & \multicolumn{1}{c|}{70.83}          & 12.384                                          \\
                          & \textbf{Ours}           & \textbf{3.753}   & \textbf{0.854}  & \multicolumn{1}{c|}{\textbf{1191}} & \textbf{45.67} & \textbf{46.53} & \multicolumn{1}{c|}{\textbf{45.73}} & \textbf{60.80} & \textbf{58.33} & \multicolumn{1}{c|}{\textbf{62.23}} & \textbf{2.839}                                  \\ \midrule
\multirow{4}{*}{Airplane} & IM-GAN                  & 3.736            & 1.110           & \multicolumn{1}{c|}{4939}          & 44.25          & 37.08          & \multicolumn{1}{c|}{\textbf{45.86}} & 79.48          & 82.94          & \multicolumn{1}{c|}{79.11}          & 21.151                                          \\
                          & SDF-StyleGAN            & 4.558            & 1.180           & \multicolumn{1}{c|}{5326}          & 40.67          & 32.63          & \multicolumn{1}{c|}{38.20}          & 85.48          & 87.08          & \multicolumn{1}{c|}{84.73}          & 26.304                                          \\
                          & MeshDiffusion           & \textbf{3.612}   & 1.042           & \multicolumn{1}{c|}{4538}          & 47.34          & 42.15          & \multicolumn{1}{c|}{45.36}          & 66.44          & 76.26          & \multicolumn{1}{c|}{\textbf{67.24}} & 11.366                                          \\
                          & \textbf{Ours}           & 3.960            & \textbf{0.902}  & \multicolumn{1}{c|}{\textbf{3167}} & \textbf{48.33} & \textbf{50.06} & \multicolumn{1}{c|}{44.13}          & \textbf{60.75} & \textbf{56.74} & \multicolumn{1}{c|}{69.16}          & \textbf{7.020}                                  \\ \midrule
\multirow{4}{*}{Rifle}    & IM-GAN                  & 3.550            & 1.058           & \multicolumn{1}{c|}{6240}          & 46.53          & 37.89          & \multicolumn{1}{c|}{42.32}          & 70.00          & 72.74          & \multicolumn{1}{c|}{69.26}          & 25.704                                          \\
                          & SDF-StyleGAN            & 4.100            & 1.069           & \multicolumn{1}{c|}{6475}          & 46.53          & 40.21          & \multicolumn{1}{c|}{41.47}          & 73.68          & 73.16          & \multicolumn{1}{c|}{76.84}          & 33.624                                          \\
                          & MeshDiffusion           & \textbf{3.124}   & 1.018           & \multicolumn{1}{c|}{5951}          & \textbf{52.63} & 42.11          & \multicolumn{1}{c|}{48.84}          & 57.68          & 67.79          & \multicolumn{1}{c|}{\textbf{55.58}} & 19.353                                          \\
                          & \textbf{Ours}           & 3.530            & \textbf{0.849}  & \multicolumn{1}{c|}{\textbf{3493}} & 48.42          & \textbf{51.58} & \multicolumn{1}{c|}{\textbf{50.53}} & \textbf{56.63} & \textbf{55.05} & \multicolumn{1}{c|}{\textbf{55.58}} & \textbf{10.951}                                 \\ \midrule
\multirow{4}{*}{Table}    & IM-GAN                  & \textbf{11.378}  & 1.567           & \multicolumn{1}{c|}{3400}          & \textbf{51.04} & 49.20          & \multicolumn{1}{c|}{51.04}          & 65.96          & 63.17          & \multicolumn{1}{c|}{62.49}          & 4.865                                           \\
                          & SDF-StyleGAN            & 13.896           & 1.615           & \multicolumn{1}{c|}{3423}          & 42.21          & 41.80          & \multicolumn{1}{c|}{42.98}          & 68.35          & 68.21          & \multicolumn{1}{c|}{66.19}          & 4.603                                           \\
                          & MeshDiffusion           & 11.405           & \textbf{1.548}  & \multicolumn{1}{c|}{3427}          & 49.56          & 50.33          & \multicolumn{1}{c|}{\textbf{51.92}} & \textbf{59.35} & 59.47          & \multicolumn{1}{c|}{\textbf{58.97}} & 4.310                                           \\
                          & \textbf{Ours}           & 11.998           & 1.564           & \multicolumn{1}{c|}{\textbf{2683}} & 46.36          & \textbf{50.41} & \multicolumn{1}{c|}{47.12}          & 61.46          & \textbf{59.43} & \multicolumn{1}{c|}{60.75}          & \textbf{3.822}                                  \\ \bottomrule
\end{tabular}
}
\label{table: metric}
\end{table*}
\begin{table*}[t]
\centering

\caption{
Quantitative evaluation of shape generation in $128^3$ resolution on airplane category. 
}

\resizebox{0.76\linewidth}{!}{
\begin{tabular}{@{}l|cccccccccc@{}}
\toprule
\multirow{2}{*}{Method} & \multicolumn{3}{c}{MMD ($\downarrow$)}                                  & \multicolumn{3}{c}{COV ($\%, \uparrow$)}                              & \multicolumn{3}{c}{1-NNA ($\%, \downarrow$)}                          & \multirow{2}{*}{JSD $(10^{-3},\ \downarrow)$} \\ \cmidrule(lr){2-10}
                        & CD $\times 10^3$ & EMD $\times 10$ & \multicolumn{1}{c|}{LFD}           & CD             & EMD            & \multicolumn{1}{c|}{LFD}            & CD             & EMD            & \multicolumn{1}{c|}{LFD}            &                                               \\ \midrule
LAS-Diffusion           & 4.654            & 0.56            & \multicolumn{1}{c|}{3142}          & 37.45          & 35.72          & \multicolumn{1}{c|}{42.15}          & 79.48          & 84.67          & \multicolumn{1}{c|}{71.51}          & 33.137                                        \\
\textbf{Ours}           & \textbf{4.000}   & \textbf{0.509}  & \multicolumn{1}{c|}{\textbf{3077}} & \textbf{46.72} & \textbf{43.88} & \multicolumn{1}{c|}{\textbf{42.27}} & \textbf{60.01} & \textbf{61.06} & \multicolumn{1}{c|}{\textbf{69.22}} & \textbf{6.873}                                \\ \bottomrule
\end{tabular}
}
\label{table: 128res}
\end{table*}
\begin{figure*}[t]
    \centering
    \includegraphics[width=0.85\linewidth]{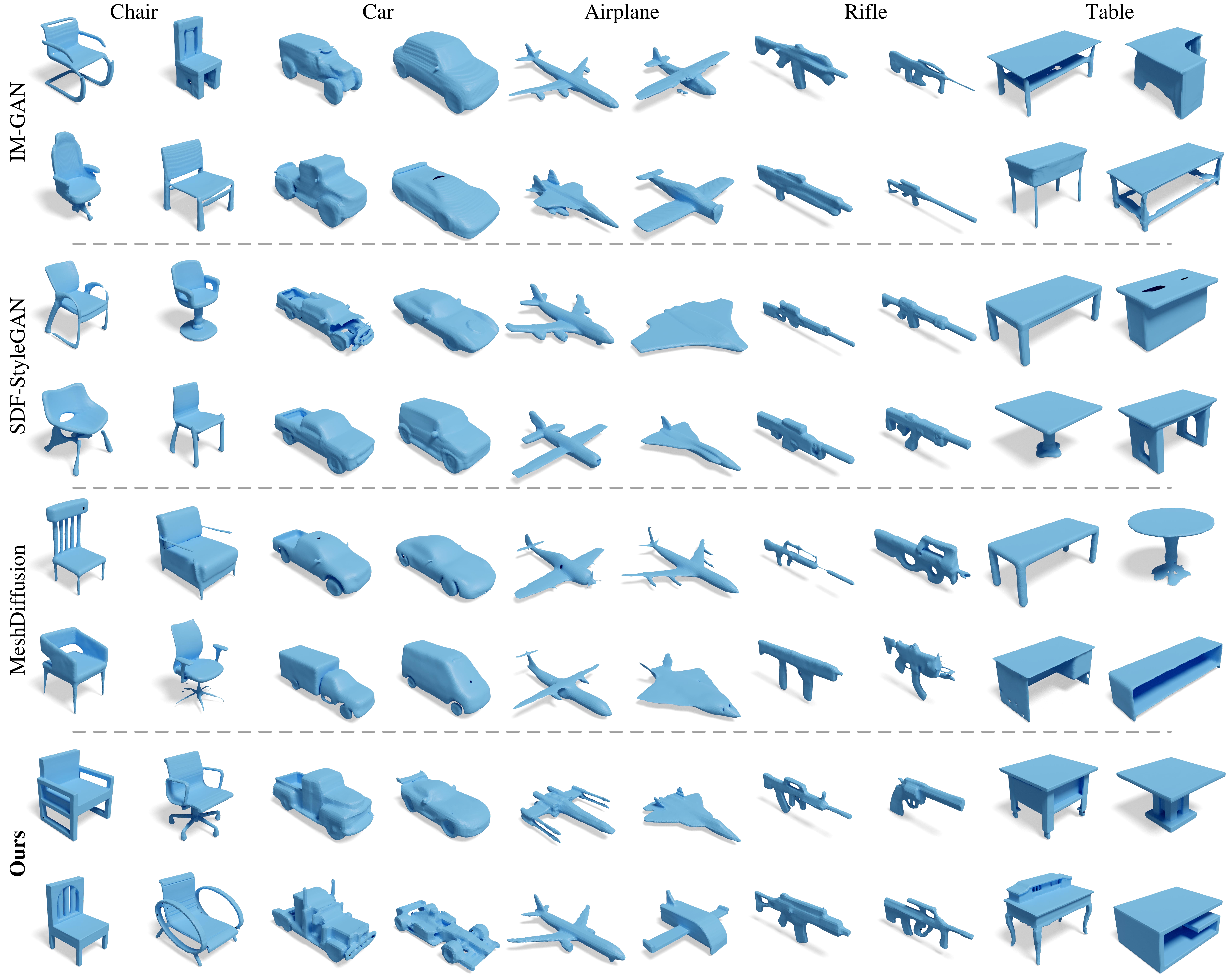}

    \caption{
    Qualitative evaluation of shape generation in $64^3$ resolution.
    }
    \label{fig: visual}
\end{figure*}

To evaluate the quality of shape generation, we compare our method with IM-GAN \cite{im_gan}, SDF-StyleGAN \cite{sdfstylegan}, MeshDiffusion \cite{meshdiffusion} and, LAS-Diffusion \cite{lasdiffusion}. 
IM-GAN, SDF-StyleGAN, and LAS-Diffusion are neural implicit function-based shape generation methods. 
IM-GAN predicts the occupancy values. 
Similarly, SDF-StyleGAN and LAS-Diffusion predict the SDFs. 
We apply MC to create meshes from synthesized implicit representations, following their protocols. 
MeshDiffusion is a mesh generation method that uses a deformable tetrahedral grid and SDF values to generate meshes directly. 
We do not compare ours with PolyGen, MeshGPT, and PolyDiff because it is unfair that their faces are limited to no more than 2800, and they cannot produce complex geometric shapes.

Four metrics and three kinds of distances are used in the quantitative experiments (\cref{table: metric}). 
We take the test dataset as the reference set $\mathcal{B}$ and generate samples $\mathcal{A}$ of the same number, i.e. $|\mathcal{A}|=|\mathcal{B}|$. 
To calculate chamfer distance (COV) and earth mover’s distance (EMD), we sample 2048 points for each mesh of $\mathcal{A}$ and $\mathcal{B}$. 
Note that all point clouds are normalized to [-1.0, 1.0], and meshes are normalized to [-0.5, 0.5]. The supplementary materials elaborate on more details of metrics.
\paragraph{Quantitative evaluation.} 
We present metric values in \cref{table: metric}. 
Our method outperforms others in most cases, indicating that our approach is superior in quality, diversity, and distribution. 
Particularly in the car and airplane category, our method performs significantly better than others.
It can be attributed to our excellent ability to generate details, considering the minimal intra-class variation within cars.
We also achieve good performance in high resolution, shown in \cref{table: 128res}.
% 记得提到实验设置与64的相同和不同
% 我们的实验指标之所以很好，在于我们的多样性很好。
% 记得提到MeshDiffusion 是在CD上训练的，因此很多时候它被偏爱。
% MMD代表参考集到生成集的最近距离的平均值。COV大致上反应了生成结果的多样性。LFD通过多角度的投影，来衡量两个东西的相似性。1-NNA衡量两个分布的一致性，越接近50%，两个分布越一致。JSD也是衡量其分布的。
% \TODO{rewrite.}
\paragraph{Qualitative evaluation.} 
We show rendered meshes of various methods in \cref{fig: visual}. 
As seen, neural implicit function-based methods tend to produce over-smooth shapes and inaccurate parts, e.g., arms of chairs, wheels of cars, and legs of tables. 
MeshDiffusion usually produces pits on surfaces due to the ambiguity and inaccurate 2D supervision, which we have examined the reason in \cref{sec: RW_shape_generation}. 
The Laplacian smoothing used in MeshDiffusion reduces its generation quality by removing details and thin parts, such as chair arms and legs, rifle barrels, aero engines, and airplane propellers. 
In contrast, our GenUDC can generate high-quality meshes with realistic appearances, various structures, and rich details.
We provide more visual samples in the supplementary materials.
% 记得提到实验设置与64的相同和不同
% 接下来就是要根据表示来解释了。
% 注意，meshdiffusion使用了chafer distance loss来，所以CD上它必然会比我们的效果好很多， tend to be favored. Our method, however, does not use CD or EMD during training.

\subsection{Comparison with NDC} 

\begin{table*}[t]
\centering

\caption{
Quantitative comparison between MC, NDC, UNDC, and ours in $64^3$ resolution on the airplane category. 
We apply those three methods to the same SDFs generated by SDF-StyleGAN. A post-processing step described in \cite{ndc} is used after UNDC.
% 记得写清楚我们在什么分辨率、什么类别下，与什么方法进行比较，为什么进行这样的间接比较（间接比较这部分可以写在正文）
}

\resizebox{0.75\linewidth}{!}{
\begin{tabular}{@{}l|cccccccccc@{}}
\toprule
                         & \multicolumn{3}{c}{MMD ($\downarrow$)}                                                                         & \multicolumn{3}{c}{COV ($\%, \uparrow$)}                                                                        & \multicolumn{3}{c}{1-NNA ($\%, \downarrow$)}                                                                    &                                                \\ \cmidrule(lr){2-10}
\multirow{-2}{*}{Method} & CD $\times 10^3$             & EMD $\times 10$              & \multicolumn{1}{c|}{LFD}                         & CD                           & EMD                          & \multicolumn{1}{c|}{LFD}                          & CD                           & EMD                          & \multicolumn{1}{c|}{LFD}                          & \multirow{-2}{*}{JSD $(10^{-3},\ \downarrow)$} \\ \midrule
SDF-StyleGAN + MC        & {\color[HTML]{333333} 4.459} & {\color[HTML]{333333} 1.113} & \multicolumn{1}{c|}{{\color[HTML]{333333} 3731}} & {\color[HTML]{333333} 41.29} & {\color[HTML]{333333} 43.88} & \multicolumn{1}{c|}{{\color[HTML]{333333} 41.14}} & {\color[HTML]{333333} 81.33} & {\color[HTML]{333333} 76.89} & \multicolumn{1}{c|}{{\color[HTML]{333333} 80.04}} & {\color[HTML]{333333} 20.581}                  \\
SDF-StyleGAN + NDC       & {\color[HTML]{333333} 7.341} & {\color[HTML]{333333} 1.257} & \multicolumn{1}{c|}{{\color[HTML]{333333} 3748}} & {\color[HTML]{333333} 17.55} & {\color[HTML]{333333} 20.64} & \multicolumn{1}{c|}{{\color[HTML]{333333} 44.00}} & {\color[HTML]{333333} 94.13} & {\color[HTML]{333333} 95.30} & \multicolumn{1}{c|}{{\color[HTML]{333333} 78.06}} & {\color[HTML]{333333} 133.827}                 \\
SDF-StyleGAN + UNDC      & {\color[HTML]{333333} 7.758} & {\color[HTML]{333333} 1.563} & \multicolumn{1}{c|}{{\color[HTML]{333333} 3902}} & {\color[HTML]{333333} 15.57} & {\color[HTML]{333333} 14.46} & \multicolumn{1}{c|}{{\color[HTML]{333333} 41.90}} & {\color[HTML]{333333} 95.24} & {\color[HTML]{333333} 97.71} & \multicolumn{1}{c|}{{\color[HTML]{333333} 80.66}} & {\color[HTML]{333333} 173.030}                 \\
\textbf{Ours}            & \textbf{3.960}               & \textbf{0.902}               & \multicolumn{1}{c|}{\textbf{3167}}               & \textbf{48.33}               & \textbf{50.06}               & \multicolumn{1}{c|}{\textbf{44.13}}               & \textbf{60.75}               & \textbf{56.74}               & \multicolumn{1}{c|}{\textbf{69.16}}               & \textbf{7.020}                                 \\ \bottomrule
\end{tabular}
}
\label{table: comparision_ndc}
\end{table*}
\begin{figure}
  % \fbox{\rule{0pt}{2in} \rule{\textwidth}{0pt}}
  % \includesvg[width=0.8\linewidth]{img/sdf_distribution.svg}
  \includegraphics[width=0.8\linewidth]{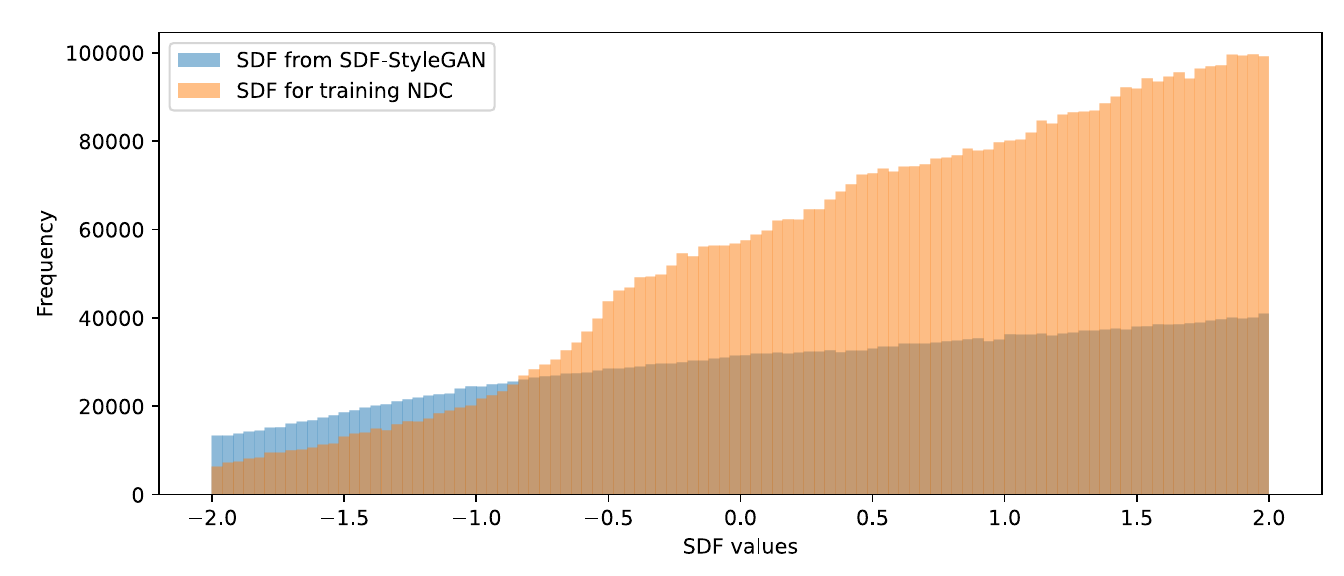}
  \caption{
  The histogram of SDFs generated by SDF-StyleGAN and SDFs for training NDC. We select 809 SDF grids and only consider SDFs near the surfaces to draw this histogram.
  }
  \label{fig: sdf_distribution}
\end{figure}
\begin{figure}[]
    \centering
    \includegraphics[width=0.8\linewidth]{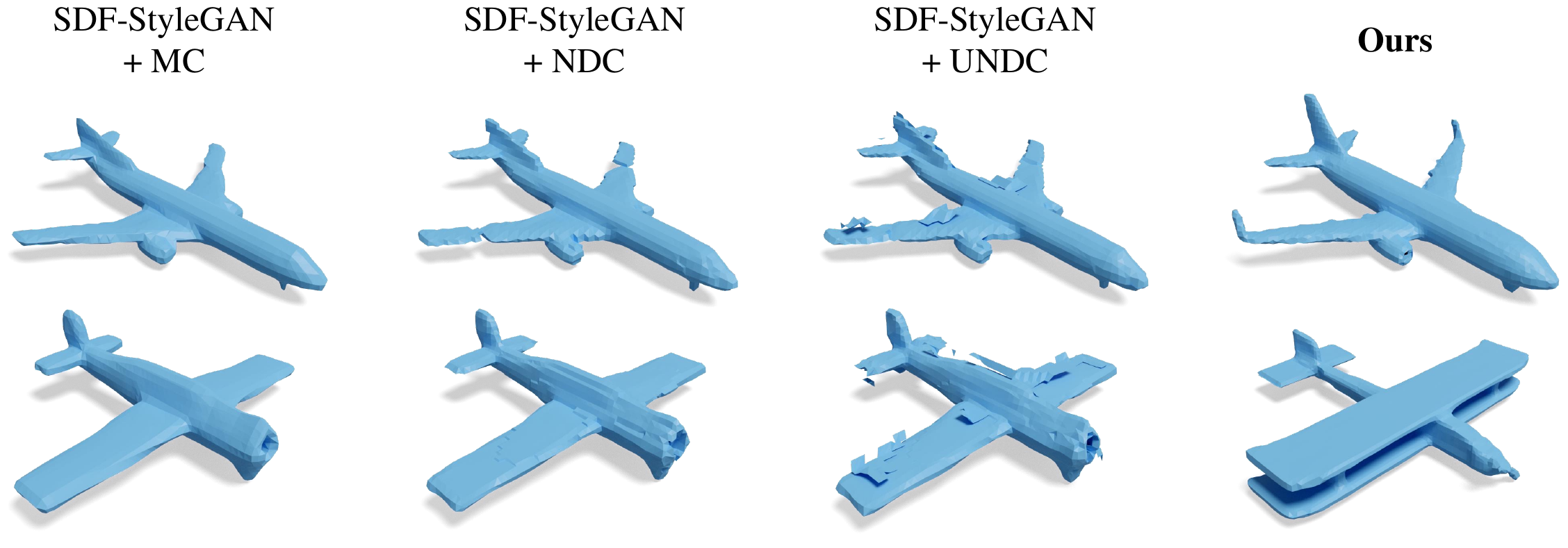}

    \caption{Visual samples of three post-processing methods and ours. We apply those post-processing methods on the same SDFs generated by SDF-StyleGAN.}
    \label{fig: comparison_with_ndc}
\end{figure}

Since NDC \cite{ndc} is an isosurface reconstruction method, we cannot directly compare GenUDC with NDC.
Thus, we train NDC and UNDC networks with the default setting of their codes and the data from the airplane category of ShapeNetCore (v1) \cite{shapenet2015} for 2500 epochs. After training, we apply them to SDFs to create meshes for comparison. 
Qualitative evaluations are shown in \cref{table: comparision_ndc}.
The performance of NDC and UDNC is poor due to the distribution gap between SDFs generated by SDF-StyleGAN and SDFs for training NDC, shown in \cref{fig: sdf_distribution}.
\cref{fig: comparison_with_ndc} shows that NDC and UDNC cannot handle the generated SDFs, resulting in surface distortion and floating artifacts.
% 这里再结合视觉效果，说说NDC和UNDC不适合的原因。
Overall, integrating NDC (UNDC) into the SDF generation method introduces too many uncertainties, making it unsuitable for mesh generation.
In contrast, our GenUDC directly generates high-quality meshes using UDC, demonstrating that our paradigm is more suitable for mesh generation.
% 这里得从指标和视觉两个角度，来阐释。

\subsection{Ablation Study of the Vertex Part Generation}
\label{sec: ablation}

\begin{figure}[]
    \centering
    \includegraphics[width=0.8\linewidth]{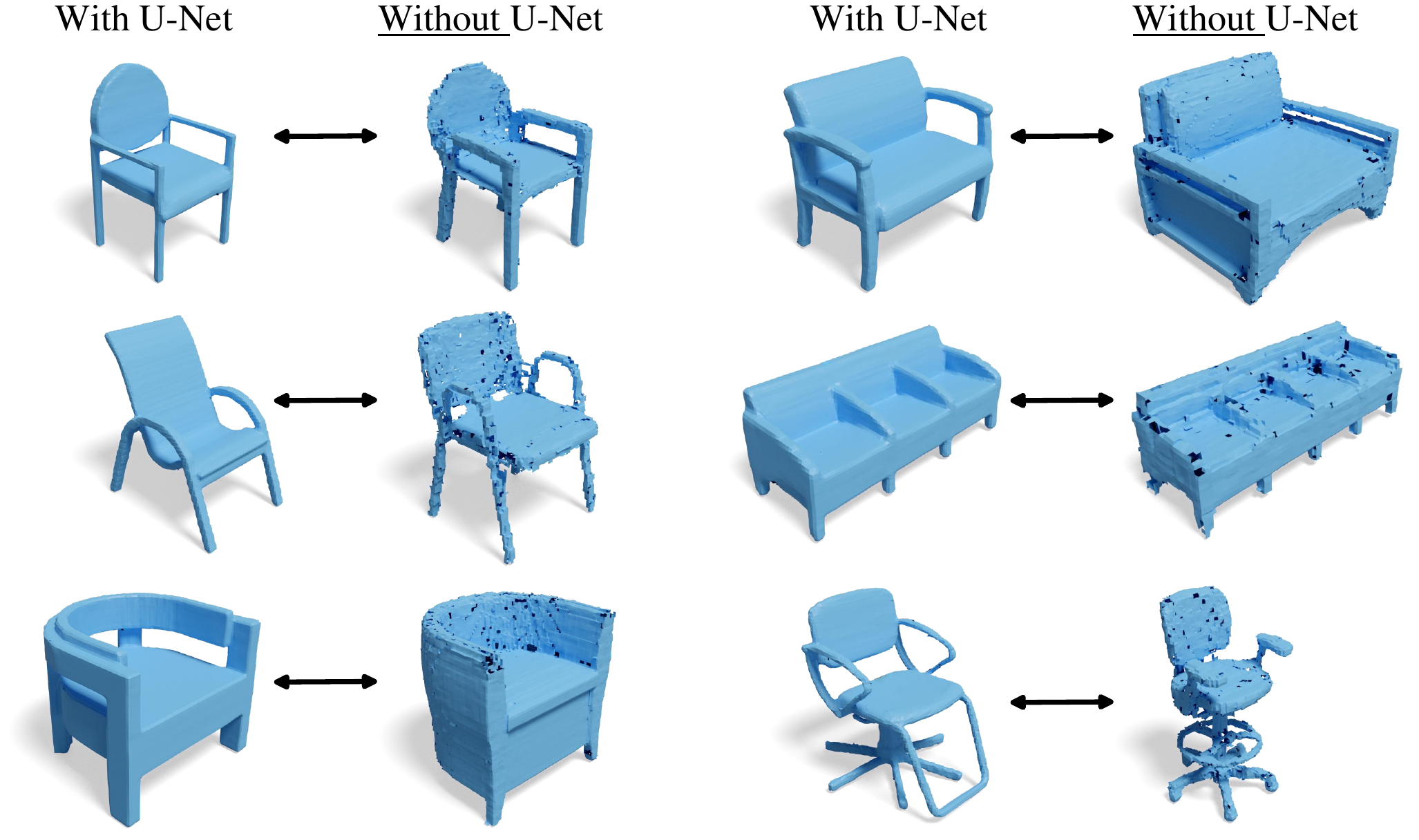}

    \caption{GenUDC with U-Net vs. GenUDC without U-Net. A pair of samples are not the same object but are similar in appearance and structure.}
    \label{fig: visual_withU_withoutU}
\end{figure}
\begin{table*}[t]
\centering

\caption{
Quantitative evaluation of ablation study. 
We compare two methods on the car category following the setting in \cref{sec: shape_generation}.
}

\resizebox{0.75\linewidth}{!}{
\begin{tabular}{@{}l|cccccccccc@{}}
\toprule
\multirow{2}{*}{Method} & \multicolumn{3}{c}{MMD ($\downarrow$)}                                  & \multicolumn{3}{c}{COV ($\%, \uparrow$)}                              & \multicolumn{3}{c}{1-NNA ($\%, \downarrow$)}                          & \multirow{2}{*}{JSD $(10^{-3},\ \downarrow)$} \\ \cmidrule(lr){2-10}
                        & CD $\times 10^3$ & EMD $\times 10$ & \multicolumn{1}{c|}{LFD}           & CD             & EMD            & \multicolumn{1}{c|}{LFD}            & CD             & EMD            & \multicolumn{1}{c|}{LFD}            &                                               \\ \midrule
Ours w/o U-net          & 15.463           & 1.702           & \multicolumn{1}{c|}{3073}          & 36.28          & 42.77          & \multicolumn{1}{c|}{36.06}          & 74.93          & 72.23          & \multicolumn{1}{c|}{75.18}          & 6.574                                         \\
Ours w/ U-net           & \textbf{14.083}  & \textbf{1.653}  & \multicolumn{1}{c|}{\textbf{2924}} & \textbf{48.08} & \textbf{48.60} & \multicolumn{1}{c|}{\textbf{47.94}} & \textbf{59.18} & \textbf{58.67} & \multicolumn{1}{c|}{\textbf{60.84}} & \textbf{4.837}                                \\ \bottomrule
\end{tabular}
}

\label{table: ablation}
\end{table*}

In this section, we compare the GenUDC to the one without the U-Net to demonstrate the necessity of the vertex refiner, i.e., U-Net. 
In the one without the U-Net, we concatenate the face part $\mathcal{F}$ and vertex part $\mathcal{V}$ as a mesh tensor and then use the LDM to learn the distribution of mesh tensors.
Other settings are consistent with the vanilla GenUDC. 
More network details are in the supplementary materials.
Then, we take mesh tensors to train the LDM, learning the joint distribution of $\mathcal{F}$ and $\mathcal{V}$.
At the inference, it simultaneously generates $\mathcal{F}$ and $\mathcal{V}$. 
However, it is quite difficult for a single LDM to learn this joint distribution and build the correlation between $\mathcal{F}$ and $\mathcal{V}$. 
To prove this, we present some similar samples produced by GenUDC with and without U-Net in \cref{fig: visual_withU_withoutU}. 
As we can see, removing U-Net results in jagged edges and unsmooth surfaces. 
% There is a relation between faces and vertices ensuring the good quality of meshes, but the LDM has difficulties learning the relation. 
Only by modeling the vertex part generation conditioned on the face part, we can learn the correlation between $\mathcal{F}$ and $\mathcal{V}$. 
The quantitative evaluation in \cref{table: ablation} also proves our opinion. % 因为face part和vertex part没有建立联系，vertex part没有考虑当前拓扑结构，所以vertex part部分的变差很容易导致锯齿。vertex part部分的生成，本质上是一个条件生成任务，vertex part部分高度依赖于face part部分。 

% \begin{figure}[t]
%     \centering
%     % \fbox{\rule{0pt}{2in} \rule{0.9\linewidth}{0pt}}
%     \includegraphics[width=\linewidth]{img/GenUDC_noUnet.pdf}

%     \caption{The GenUDC without the U-Net for vertices generation. Since we remove the U-Net of vertex part generation, the VAE reconstructs the face part $\mathcal{F}$ and vertex part $\mathcal{V}$ together instead of only reconstructing $\mathcal{F}$.}
%     \label{fig: framework_noUnet}
% \end{figure}

\subsection{Data Fitting Comparison}
\label{sec: data_fiting}

Our UDC is a discretized mesh counterpart, which requires a data fitting process. 
In this section, we demonstrate the superiority of UDC in the data fitting process compared with MeshDiffusion, which uses a deformable tetrahedral grid to discretize a mesh.

For quantitative evaluation, we randomly select one hundred meshes and record the average processing time and memory footprints in \cref{table: data_fitting}. 
% We compare our method with MeshDiffusion \cite{meshdiffusion}, which takes SDF values and a deformable grid as the mesh representation. 
As shown, UDC outperforms MeshDiffusion in both speed and memory footprint. 
The reason is that MeshDiffusion uses the rendered 2D images as the supervision of data fitting. 
Rendering 2D images requires a lot of GPU and CPU resources, and it takes a long time to fit data. 
In contrast, we only use the CPU to directly calculate the fitting vertices and faces of UDC as we elaborate in \cref{sec: method_udc}, which is resource-efficient and fast.  
% 我们的方法在什么上有优势，为什么有优势 
% 平均值、中位值、最大值、最小值、平均GPU、平均CPU消耗。说明一下上述结果是在什么CPU、GPU、内存上运行的。然后根据上面的不同值，介绍下我们方法的情况。

\begin{table}[t]
\centering

\caption{
Quantitative evaluation of data fitting in terms of mean processing time, GPU memory footprints, and CPU memory footprints of one hundred samples. 
All the programs are executed in a single thread, using an NVIDIA RTX 3090 GPU, an Intel i7-10700 CPU, and 64GB of memory.
}

\resizebox{0.75\linewidth}{!}{
\begin{tabular}{@{}c|ccc@{}}
\toprule
                    & \begin{tabular}[c]{@{}c@{}}Processing time \\ (Sec.)\end{tabular} & \begin{tabular}[c]{@{}c@{}}GPU memory \\ (MB)\end{tabular} & \begin{tabular}[c]{@{}c@{}}CPU memory \\ (MB)\end{tabular} \\ \midrule
MeshDiffusion       & 1,408                                                             & 8544                                                       & 1497                                                       \\
\textbf{Ours (UDC)} & \textbf{43}                                                       & \textbf{0}                                                 & \textbf{1266}                                              \\ \bottomrule
\end{tabular}
}

\label{table: data_fitting}
\end{table}
\begin{figure}[]
    \centering
    \includegraphics[width=0.8\linewidth]{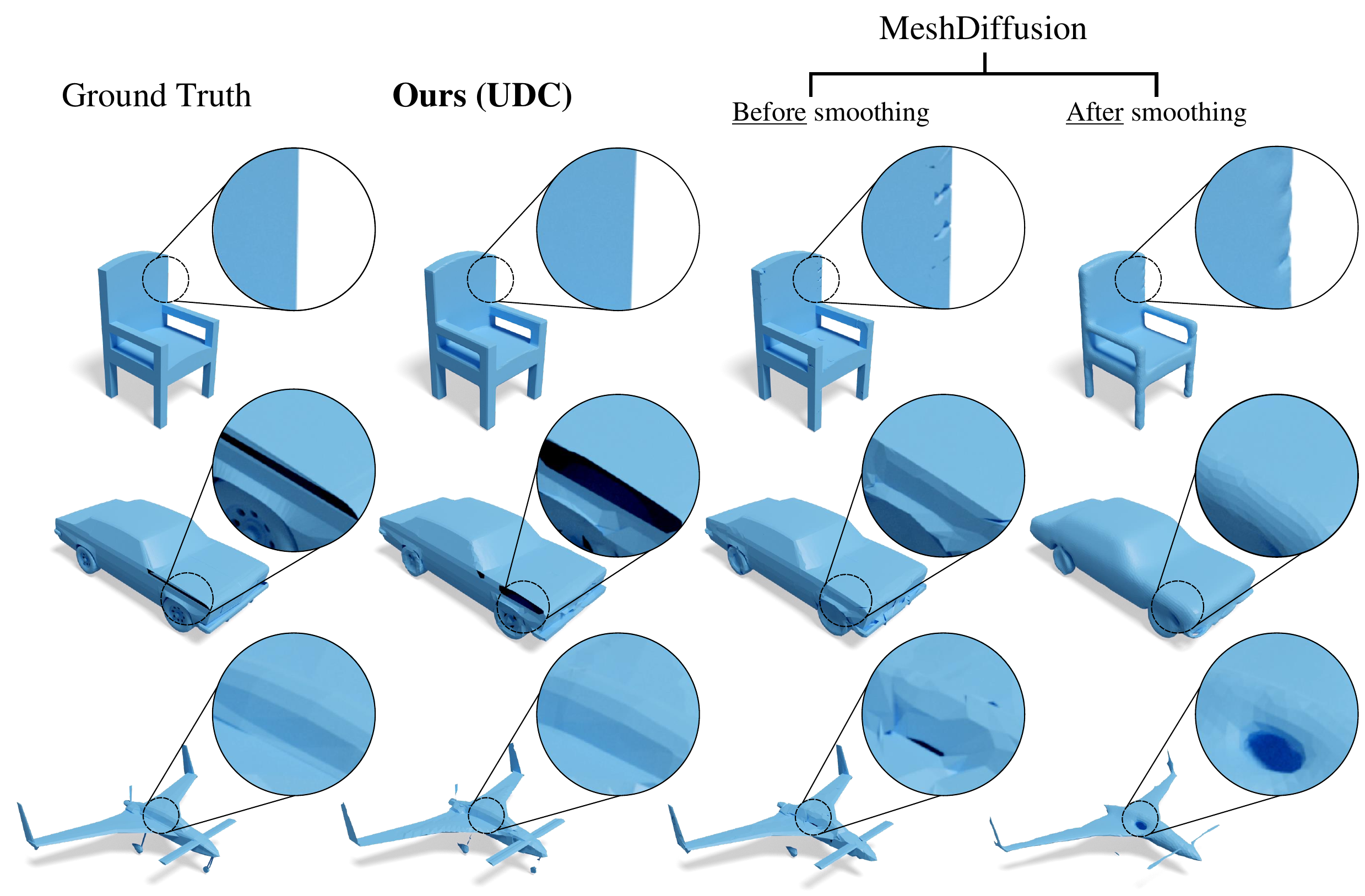}

    \caption{Qualitative evaluation of data fitting. The resolution of those meshes is $64^3$ except for the ground truth. In addition, since MeshDiffusion applies a post-processing method to the uneven surfaces of its fitting mesh, we present both raw and smoothed meshes.}
    \label{fig: data_fitting}
\end{figure}

To visually illustrate UDC's superiority, we present some samples in \cref{fig: data_fitting}.
As seen, MeshDiffusion is unavoidable to produce pits on the mesh surfaces and lack of details, such as the line and crack on the car. 
The reason is the ambiguity and inaccurate 2D supervision discussed in \cref{sec: RW_shape_generation}. 
Laplacian smoothing used by MeshDiffusion even removes details and sharp parts instead of the pits. 
In comparison, UDC can fit flat surfaces, sharp parts, and curved surfaces with details. 
% Moreover, UDC does not produce unreasonable pits on the mesh surfaces.
% 说清楚分辨率。说清楚我们甚至只是用了单线程，如果用多线程，会更快。 然后结合图，来说明我们和他们的区别，他们的缺陷、我们的优势。他们为什么会有这个缺点，因为通过单图渲染进行监督。我们为什么会有这个优势，因为我们是直接在三维层面上进行计算。

\section{Limitation}

Firstly, our main limitation is the non-manifold issue. 
Since we adopt UDC as the mesh representation, our method inherits the non-manifold issue from DC. 
However, such an issue rarely occurs. 
It can be resolved by "tunneling" through vertices/edges or dividing them with the approaches introduced by \cite{dualMC, manifoldDC}. 
Secondly, the memory footprint constrains the resolution of our results.
Thirdly, since the face part is a set of boolean values, our models may predict wrong boolean values, resulting in pits on the surface. 
We can solve this pit problem with the post-processing method in \cite{ndc}.
% 注意NDC的图16正下方的段落，However开头的那一句，有说明UNDC的可能并不适合所有的application。

\section{Conclusion \& Future Works}

% 主要结果、贡献
% 高层次地回顾正片文章的故事，更细致地讲解工作的新颖性、影响力。——我们提出了什么方法用于高质量的三维mesh生成。
In conclusion, we propose a novel 3D generative framework, GenUDC, using the Unsigned Dual Contouring representation (UDC) for high-quality mesh generation. 
Our method can directly generate high-quality meshes without using isosurface reconstruction methods.
Specifically, following the discretization idea, we fit a mesh in a regular grid to get its UDC representation.
Since UDC is composed of the face and vertex parts, we use a two-stage, coarse-to-fine pipeline to learn its distribution. 
Firstly, we use a latent diffusion model to generate the face part.
Secondly, we take a U-Net as a vertex refiner to synthesize the vertex part conditioned on the face part.
Experiments demonstrate our superiority over baselines in shape generation and data fitting.
The ablation study proves the validity of network design.
We believe that our method offers a new paradigm for further work in mesh generation.

% 对未来工作进行回顾。
In the future, we plan to apply GenUDC to various applications, such as text-to-3D, joint generation of texture and shape, single view 3D reconstruction, shape editing, 3D attacks \cite{gao2023imperceptible}, etc.

\begin{acks}
The research was sponsored by the National Natural Science Foundation of China (No. 62176170, 61773270).
\end{acks}

\bibliographystyle{ACM-Reference-Format}
\bibliography{acmmm_ref}

\newpage

\appendix

% \clearpage
\setcounter{page}{1}
% \maketitlesupplementary
\renewcommand{\thefigure}{\Alph{figure}}

\section{The Architecture of Networks}

\begin{figure}[t]
    \centering
    \includegraphics[width=0.65\linewidth]{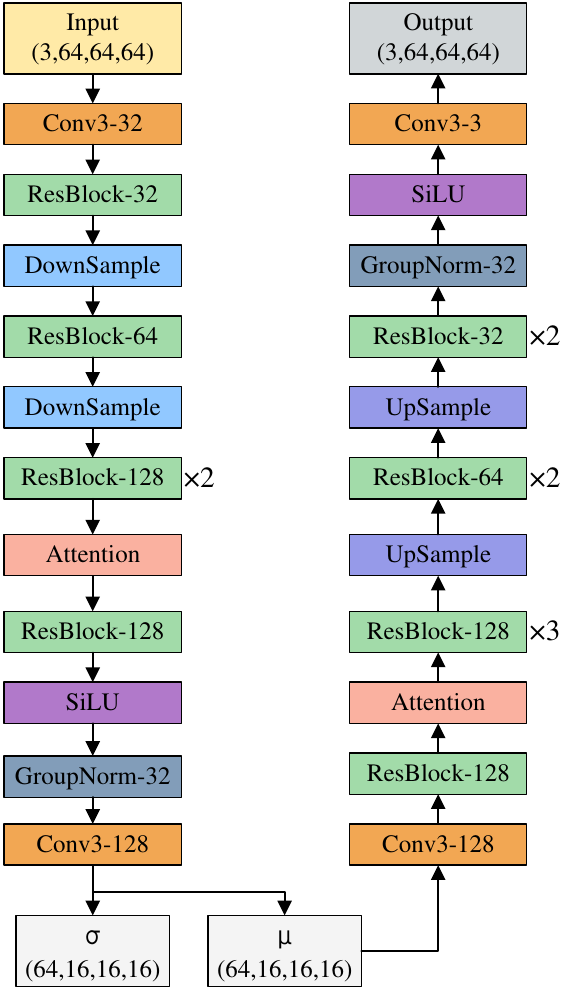}

    \caption{
    Our VAE network for 64 resolution.
    }
    \label{fig: supple_vae64}
    \Description{}
\end{figure}

\begin{figure}[t]
    \centering
    \includegraphics[width=0.65\linewidth]{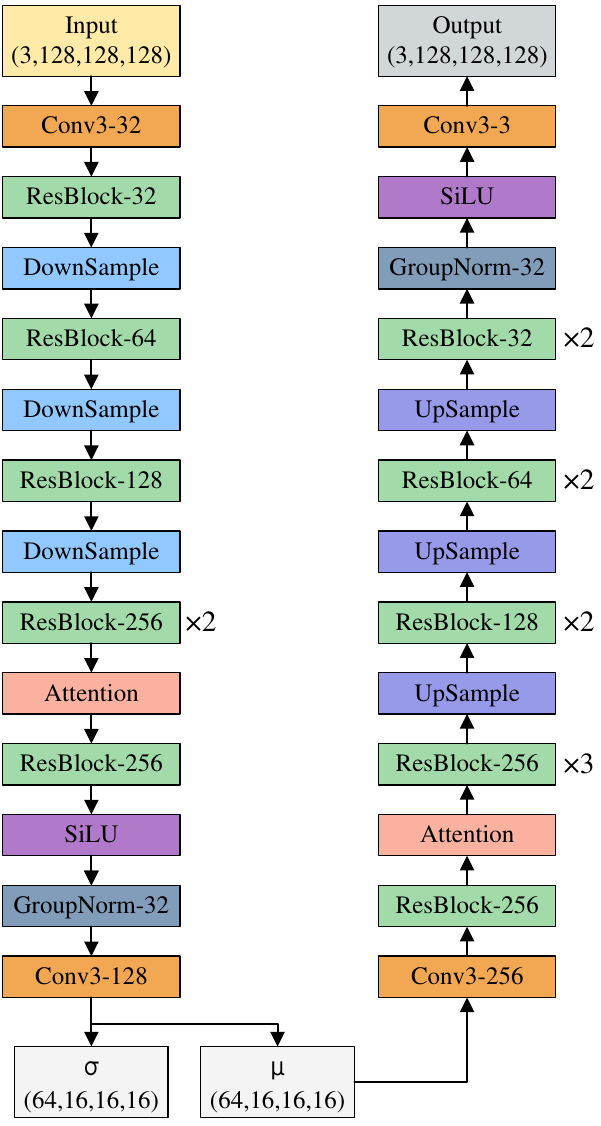}

    \caption{
    Our VAE network for 128 resolution.
    }
    \label{fig: supple_vae128}
    \Description{}
\end{figure}
\begin{figure}[t]
    \centering
    \includegraphics[width=0.7\linewidth]{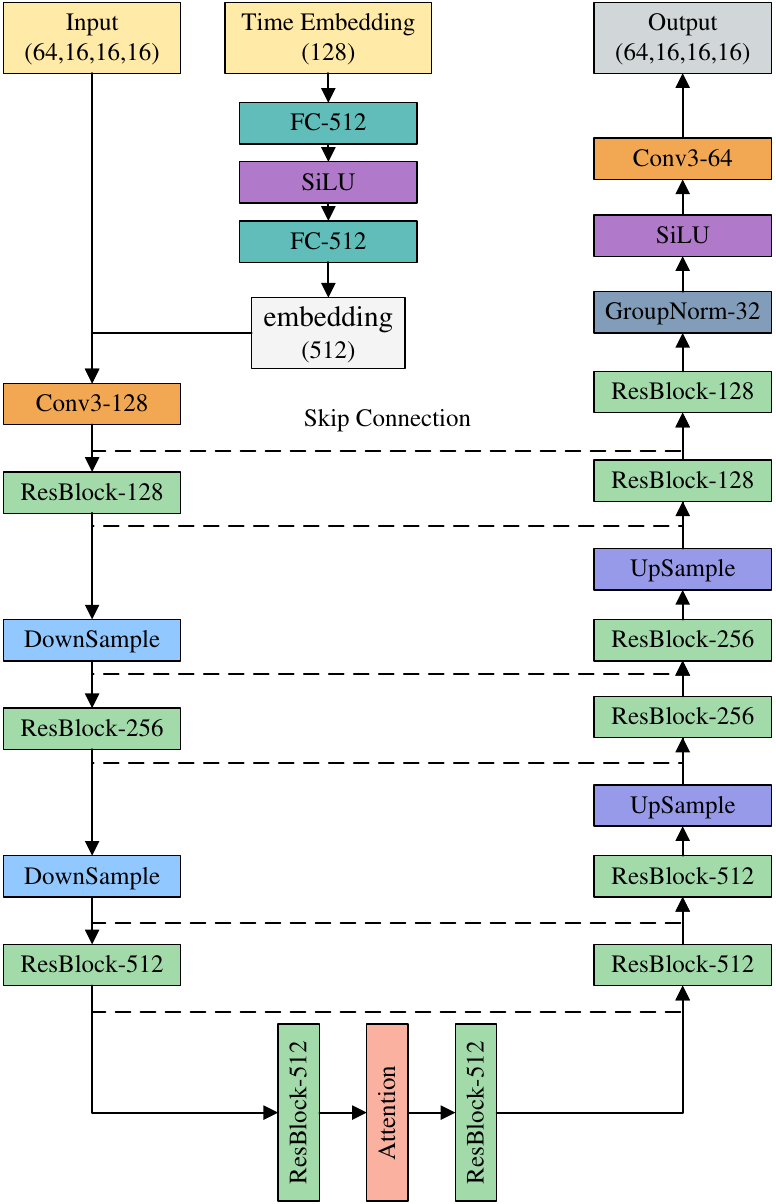}

    \caption{
    Our diffusion model for 64 and 128 resolution.
    }
    \label{fig: supple_diffusion64}
    \Description{}
\end{figure}
\begin{figure}[t]
    \centering
    \includegraphics[width=0.7\linewidth]{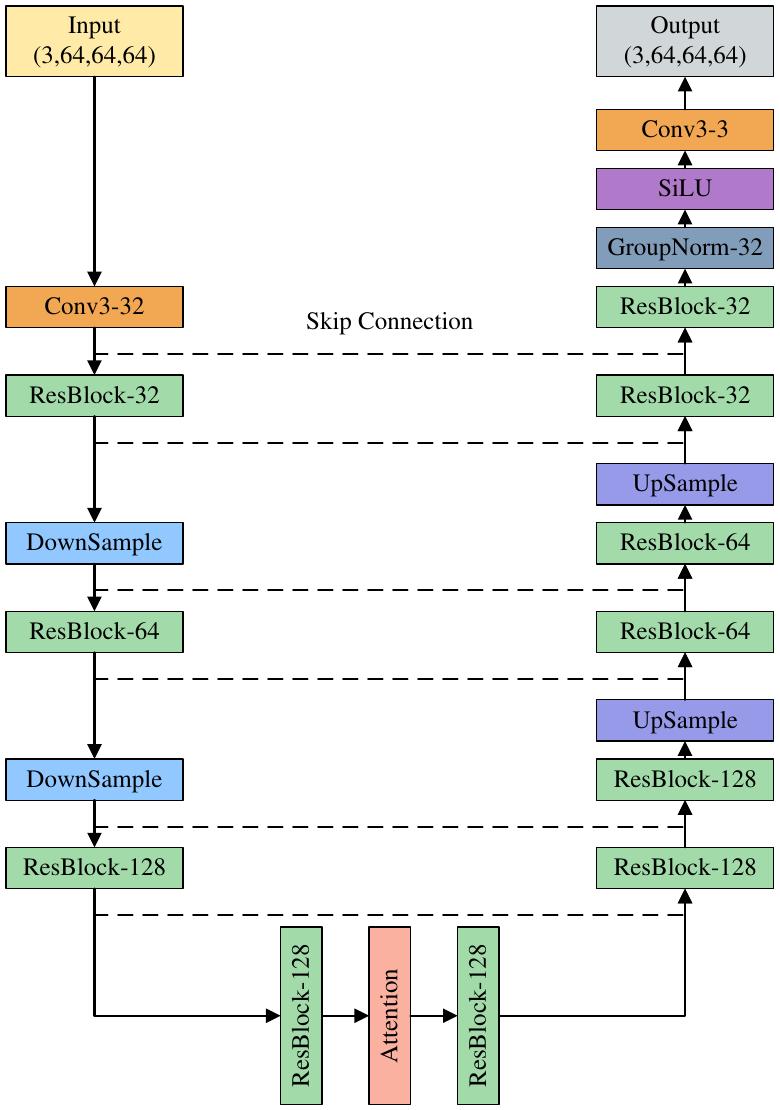}

    \caption{
    Our vertex refiner (U-Net) for 64 resolution.
    }
    \label{fig: supple_unet64}
    \Description{}
\end{figure}

\begin{figure}[t]
    \centering
    \includegraphics[width=0.7\linewidth]{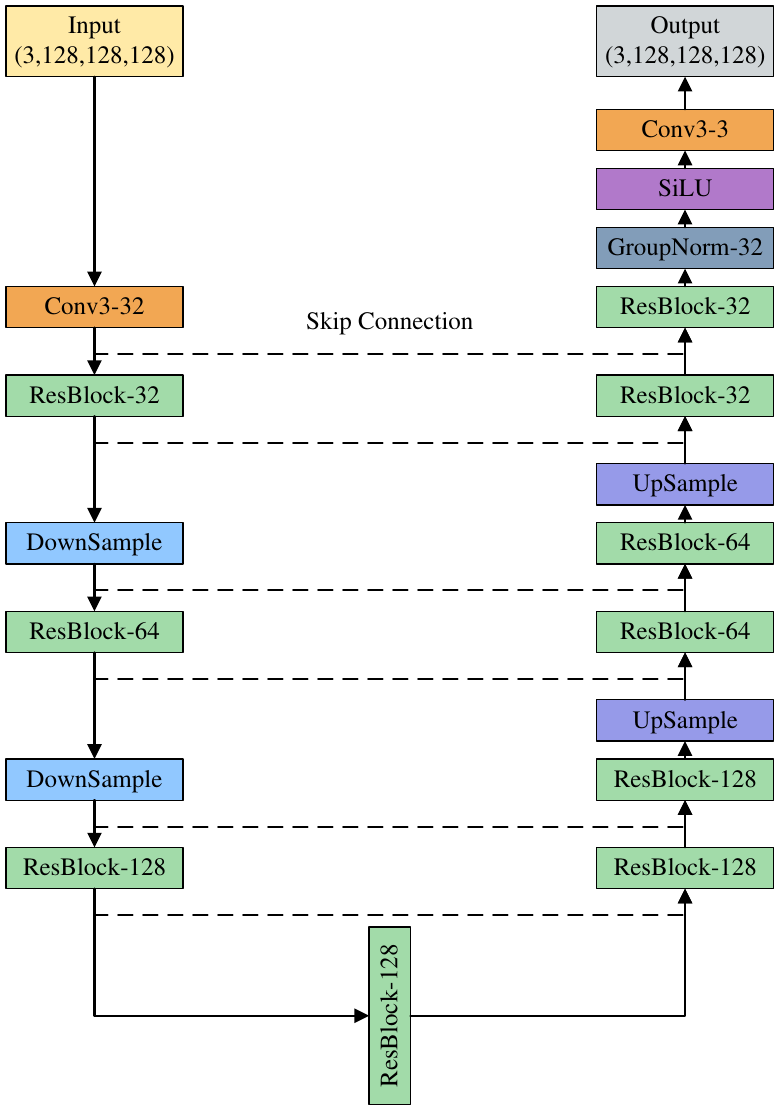}

    \caption{
    Our vertex refiner (U-Net) for 128 resolution.
    }
    \label{fig: supple_unet128}
    \Description{}
\end{figure}
\begin{figure}[t]
    \centering
    \includegraphics[width=0.5\linewidth]{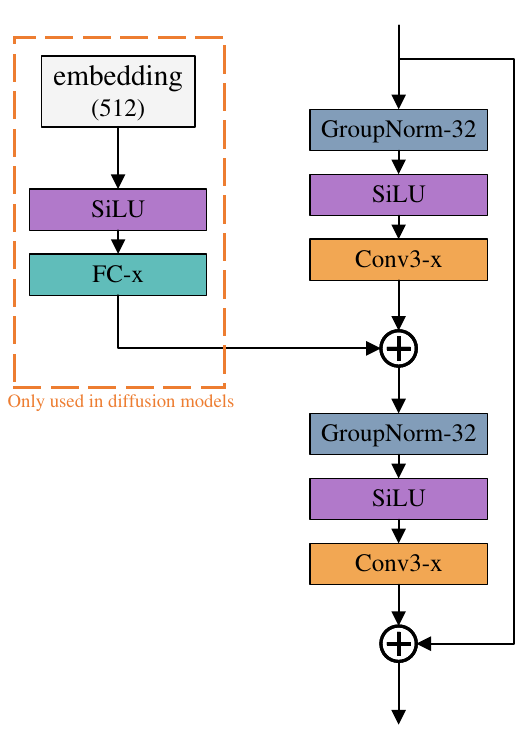}

    \caption{
    The architecture of our ResNet Block.
    }
    \label{fig: supple_resblock}
    \Description{}
\end{figure}

We show the architecture of our networks in \cref{fig: supple_vae64}, \cref{fig: supple_vae128}, \cref{fig: supple_diffusion64}, \cref{fig: supple_unet64}, and \cref{fig: supple_unet128}. 
Specifically, Conv\textit{k}-\textit{x} is the convolution layer with \textit{k} kernel size, \textit{x} output channels, 1 stride, 1 padding. 
ResBlock-\textit{x} is the ResNet block \cite{resnet} with \textit{x} output channels. 
We present the details of ResBlock-\textit{x} in \cref{fig: supple_resblock}.
SiLU is the silu function. 
GroupNorm-\textit{x} is the group normalization with \textit{x} groups. 
The scale factor of DownSample and UpSample is 2. 
Attention is the attention block.
FC-\textit{x} is the fully connected layer with \textit{x} output channels.

\section{diffusion model}
A diffusion model consists of two opposite processes: the forward process and the reverse process. Given the latent representation $z_0\sim p(z_0)$ as the data, the forward process adds the controlled Gaussian noise $\epsilon$ to $z_0$ for $T$ times:
\begin{align}
\label{eq:dm_forward}
q(z_T|z_0)&=\prod_{t=1}^{T} q(z_t|z_{t-1}), \\
q(z_t|z_{t-1})&=N(z_t;\sqrt{\alpha_t}z_{t-1},\beta_t I),
\end{align}
where $\alpha_t=1-\beta_t$, and $\beta_t$ is the predefined variance. In contrast, the reverse process denoises $z_T$ to $z_0$:
\begin{align}
\boldsymbol{p_{\theta}}(z_0|z_T)&=\prod_{t=1}^{T} \boldsymbol{p_{\theta}}(z_{t-1}|z_{t}), \\
\boldsymbol{p_{\theta}}(z_{t-1}|z_{t})&=N(z_{t-1};\boldsymbol{\mu_\theta}(z_t, t),\beta_t I).
% &\mu_\theta(x_t, t)=\frac{1}{\sqrt[]{\alpha_t}}\left(x_t-\frac{\beta_t}{\sqrt[]{1-\bar{\alpha}_t}} \epsilon_\theta(x_t, t)\right)
\end{align}
According to the method in DDPM \cite{diffusion-model}, we reparameterize $\boldsymbol{\mu_\theta}(z_t, t)$ as:
\begin{align}
\boldsymbol{\mu_\theta}(z_t, t)&=\frac{1}{\sqrt[]{\alpha_t}}\left(z_t-\frac{\beta_t}{\sqrt[]{1-\bar{\alpha}_t}} \boldsymbol{\epsilon_\theta}(z_t, t)\right), \\
\label{eq:xt}
z_t&=\sqrt{\bar{\alpha}_t}z_0+\sqrt{1-\bar{\alpha}_t}\boldsymbol{\epsilon} \text{, \ } \bar{\alpha}_t=\prod_{i=1}^{t}\alpha_i,
\end{align}
where $\boldsymbol{\epsilon}\sim \mathcal{N}(0,1)$  is the noise. In the cycle of the forward process and the reverse process, the noise $\boldsymbol{\epsilon_\theta}(z_t, t)$ is the only unknown value. Thus, we predict $\boldsymbol{\epsilon_\theta}(z_t, t)$ by a neural network parameterized as $\theta$ to complete the cycle.

We train the network $\boldsymbol{\epsilon_\theta}$ with:
\begin{align}
\mathcal{L}_{dm}=  \mathbb{E}_{z,t,\epsilon \sim \mathcal{N}(0,1)} ||\boldsymbol{\epsilon} - \boldsymbol{\epsilon_\theta}(z_t, t)||_1 .
\end{align}
In the forward process, we add the noise $\boldsymbol{\epsilon}$ to $z_0$ for getting $z_t$ as shown in \cref{eq:xt}, and train the network $\boldsymbol{\epsilon_\theta}$ to fit $\boldsymbol{\epsilon}$. After training, our U-Net denoises $z_T \sim \mathcal{N}(0,1)$ to $z_0$ in the reverse process. 
% Then the latent representation $z_0$ is decoded as the face part $\mathcal{F}$ by the pretrained decoder of VAE.

\section{Metrics}
% 四个指标，3个距离度量

\quad \textbf{Chamfer distance (CD)} measures the similarity between two point clouds. The CD is formulated as:
\begin{equation}
\begin{aligned}
    CD(A,B)&=\sum_{a \in A}^{}\min_{b \in B}||a-b||^{2}_{2} \\
    &+ \sum_{b \in B}^{}\min_{a \in A}||a-b||^{2}_{2},
\end{aligned}
\end{equation}
where $A$ and $B$ are generated point clouds and reference point clouds. 

\textbf{Earth mover’s distance (EMD)} is a metric of dissimilarity between two distributions and can be also used to measure the similarity between two point clouds:
\begin{equation}
    EMD(A,B)=\min_{\phi: A\rightarrow B}\sum_{a\in A}^{}||a-\phi(a)||_2,
\end{equation}
where $\phi$ is the bijection between $A$ and $B$.

\textbf{Light field descriptor (LFD)} \cite{im_gan, grid-im-gan, lfd} utilize silhouette images rendered from 20 camera poses to measure the structure similarity between two shapes.

\textbf{Jensen-shannon divergence (JSD)} is calculated between two marginal point distribution of $A$ and $B$:
\begin{equation}
\begin{aligned}
    JSD(P_X, P_Y)=&\frac{1}{2}D_{KL}(P_X||M) \\
    +&\frac{1}{2}D_{KL}(P_Y||M),
\end{aligned}
\end{equation}
where $P_X$ and $P_Y$ are marginal distributions of points in the generated point clouds $A$ and the reference point clouds $B$ respectively. To approximate, we discretize the point cloud space into $28^3$ voxels and assign each point to one of $P_X$ and $P_Y$.

\textbf{Coverage (COV)} measures the diversity of generated dataset $X$ in comparison to the reference dataset $Y$. For each $x \in X$, it finds a nearest neighbor $y \in Y$ as a match. COV is the fraction of matched $y$ in the reference dataset $Y$:
\begin{equation}
   COV(X,Y)=\frac{\left|  \{ \underset {y\in Y} { \operatorname {arg\,min}}\ D(x,y)| x\in X  \} \right|}{|Y|},
\end{equation}
where $D(\cdot,\cdot)$ is a distance function, such as CD, EMD, or LFD.

\textbf{Minimum matching distance (MMD)} measures the quality of $X$ referred to $Y$. For each $y \in Y$, it finds the nearest neighbor $x$ in $X$ and records $D(x,y)$. MMD is the mean of those distances:
\begin{equation}
    MMD(X,Y)=\frac{1}{|Y|}\sum_{y\in Y}^{}\underset{x\in X}{\arg\min}\ D(x,y)
\end{equation}

\textbf{1-nearest neighbor accuracy (1-NNA)} \cite{1nn} measures the similarity between two distributions: 
\begin{equation}
\begin{aligned}
    1\text{-}NNA(X,Y)&= \frac{\sum_{x\in X}^{}\mathbb{I}(n_x\in X)}{|X|+|Y|} \\ &+ \frac{\sum_{y\in Y}^{}\mathbb{I}(n_y\in Y)
}{|X|+|Y|},
\end{aligned}
\end{equation}
where $n_x$ is the nearest neighbor of $x$ in $X  \cup Y - \{x\}$ and $\mathbb{I}(\cdot)$ is a indicator function. For example, if $n_x\in X$, $\mathbb{I}(n_x\in X)=1$. If $n_x\notin X$, $\mathbb{I}(n_x\in X)=0$. Ideally, if $X$ and $Y$ are sampled from the same distribution, the 1-NNA value should be 50\%. The closer the 1-NNA value is to 50\%, the more similar $X$ and $Y$ are.

\section{GenUDC without U-Net}

\begin{figure}[t]
    \centering
    \includegraphics[width=\linewidth]{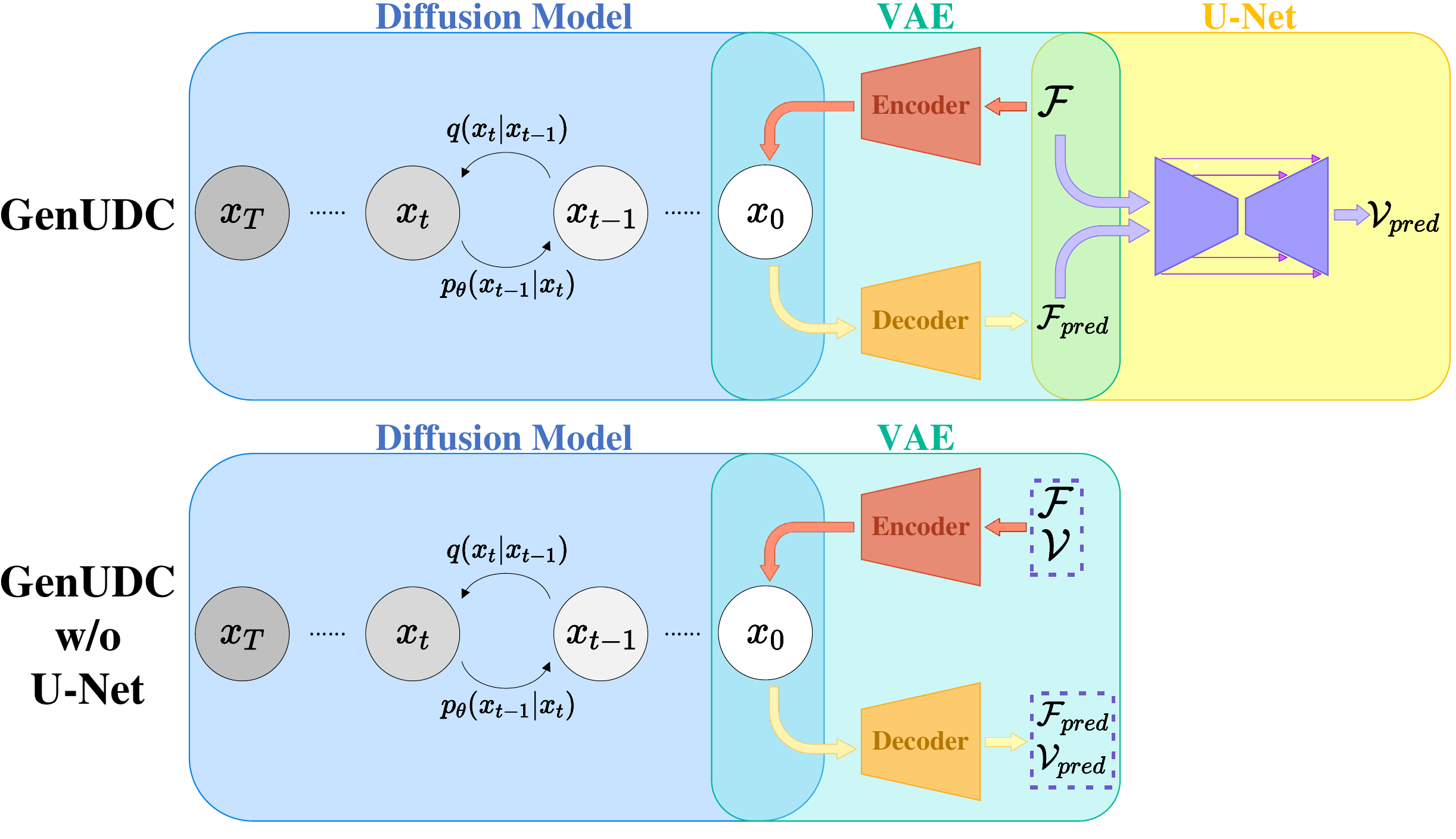}

    \caption{
    The pipeline differences between GenUDC and GenUDC without U-Net.
    }
    \label{fig: genudc_noUnet}
    \Description{}
\end{figure}

We present the pipeline of GenUDC without U-Net in \cref{fig: genudc_noUnet}. In GenUDC without U-Net, we remove the vertex refiner (U-Net) and concatenate the face part $\mathcal{F}$ and the vertex part $\mathcal{V}$ together to train the latent diffusion model.

\section{More Visual Samples of GenUDC}
\begin{figure}[t]
    \centering
    \includegraphics[width=\linewidth]{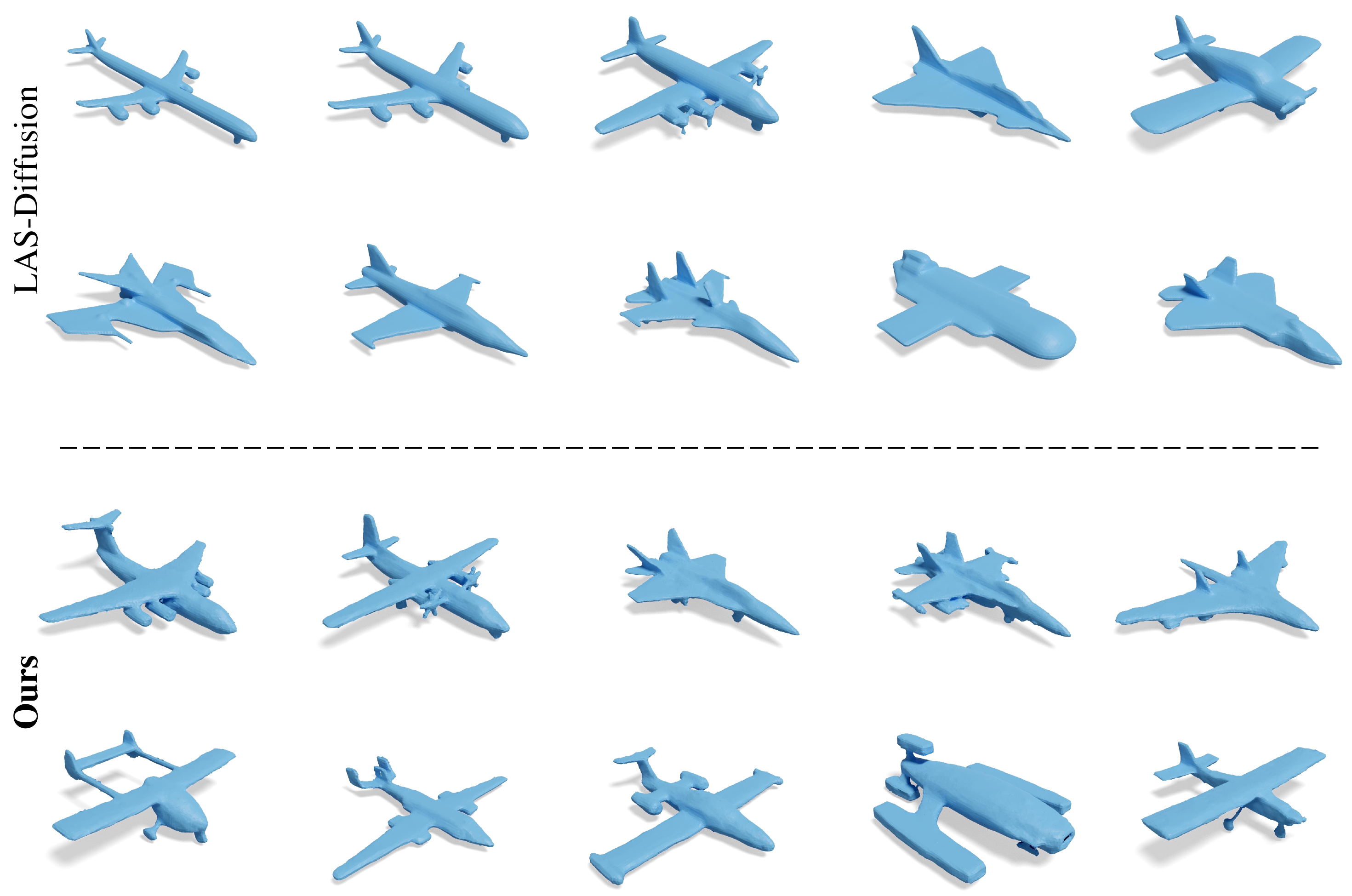}

    \caption{
    Qualitative evaluation of shape generation in $128^3$ resolution.
    }
    \label{fig: supple_128res}
    \Description{}
\end{figure}

We present visual samples of 128 resolution in \cref{fig: supple_128res}. 
Both methods are visually good. 
It is difficult to distinguish which method is better according to those visual samples.
However, Tab. 3 of the main paper proves that our variety and distribution are better than LAS-Diffusion \cite{lasdiffusion}.

% \section{Shape Novelty Analysis}
% 仿照waveley 和meshgpt的实验，来做分析，但不急。

\end{document}